\documentclass[10pt,journal,compsoc]{IEEEtran}
\ifCLASSOPTIONcompsoc
  \usepackage[nocompress]{cite}
\else
  \usepackage{cite}
\fi

\usepackage{amsmath}
\usepackage{amssymb}
\usepackage{amsfonts}
\usepackage{algorithm}
\usepackage{booktabs}
\usepackage{caption}
\usepackage{color}
\usepackage{epsfig}
\usepackage[linguistics]{forest}
\usepackage{graphicx}
\usepackage{makecell}
\usepackage{mathtools}
\usepackage{multirow,makecell}
\usepackage{physics}
\usepackage{soul}
\usepackage{url}
\usepackage{tikz}
\usetikzlibrary{matrix,shapes,arrows,fit,tikzmark}
\usepackage{subcaption}

\begin{document}
\title{Multi-Task Learning for Dense Prediction Tasks: A Survey}
%
%
%
%

\author{Simon~Vandenhende,~Stamatios~Georgoulis,~Wouter~Van~Gansbeke,\\~Marc~Proesmans,~Dengxin~Dai~and~Luc~Van~Gool
\IEEEcompsocitemizethanks{\IEEEcompsocthanksitem Simon Vandenhende, Wouter Van Gansbeke and Marc Proesmans are with the Center for Processing Speech and Images, Department Electrical Engineering, KU Leuven. E-mail: \{simon.vandenhende,wouter.vangansbeke,marc.proesmans\}@kuleuven.be\protect\\
\IEEEcompsocthanksitem Stamatios Georgoulis and Dengxin Dai are with the Computer Vision Lab, Department Electrical Engineering, ETH Zurich. E-mail: \{georgous,daid\}@ee.ethz.ch\protect\\
\IEEEcompsocthanksitem Luc Van Gool is with both the Center for Processing Speech and Images, KU Leuven and the Computer Vision Lab, ETH Zurich. E-mail: vangool@vision.ee.ethz.ch\protect\\}
\thanks{Manuscript received September X, 2020}}

\IEEEtitleabstractindextext{%
\begin{abstract}
With the advent of deep learning, many dense prediction tasks, i.e. tasks that produce pixel-level predictions, have seen significant performance improvements. The typical approach is to learn these tasks in isolation, that is, a separate neural network is trained for each individual task. Yet, recent multi-task learning (MTL) techniques have shown promising results w.r.t. performance, computations and/or memory footprint, by jointly tackling multiple tasks through a learned shared representation. In this survey, we provide a well-rounded view on state-of-the-art deep learning approaches for MTL in computer vision, explicitly emphasizing on dense prediction tasks. Our contributions concern the following. First, we consider MTL from a network architecture point-of-view. We include an extensive overview and discuss the advantages/disadvantages of recent popular MTL models. Second, we examine various optimization methods to tackle the joint learning of multiple tasks. We summarize the qualitative elements of these works and explore their commonalities and differences. Finally, we provide an extensive experimental evaluation across a variety of dense prediction benchmarks to examine the pros and cons of the different methods, including both architectural and optimization based strategies.
\end{abstract}

}

\maketitle

\IEEEdisplaynontitleabstractindextext

%
\IEEEpeerreviewmaketitle

\IEEEraisesectionheading{\section{Introduction}\label{sec:introduction}}

\IEEEPARstart{O}{ver} the last decade, neural networks have shown impressive results for a multitude of tasks, such as semantic segmentation~\cite{long2015fully}, instance segmentation~\cite{he2017mask} and monocular depth estimation~\cite{eigen2014depth}. Traditionally, these tasks are tackled in isolation, i.e. a separate neural network is trained for each task. Yet, many real-world problems are inherently multi-modal. For example, an autonomous car should
be able to segment the lane markings, detect all instances in the scene, estimate their distance and trajectory, etc., in order to safely navigate itself in its surroundings. Similarly, an intelligent advertisement system should be able to detect the presence of people in its viewpoint, understand their gender and age group, analyze their appearance, track where they are looking at, etc., in order to provide personalized content. At the same time, humans are remarkably good at solving many tasks concurrently. Biological data processing appears to follow a multi-tasking strategy too: instead of separating tasks and tackling them in isolation, different processes seem to share the same early processing layers in the brain (see V1 in macaques~\cite{gur2007direction}). The aforementioned observations have motivated researchers to develop generalized deep learning models that given an input can infer all desired task outputs.

\definecolor{level1}{RGB}{244,142,130}
\definecolor{level2}{RGB}{99,175,243}
\definecolor{level3}{RGB}{116,223,123}
\definecolor{level4}{RGB}{255,224,128}

\begin{figure*}[ht]
\centering
\medskip
\small
\begin{forest}
for tree={s sep=1mm, inner sep=2, l=1}
[
{Multi-Task Learning Methods },fill=level1
[Deep Multi-Task\\ Architectures \textbf{(Sec.~\ref{sec: deep_multi_task_architectures})},fill=level2
[Encoder-Focused\\ \textbf{(Sec.~\ref{subsec: encoder_focused_architectures})},fill=level3
[{MTL Baseline}\\Cross-Stitch Networks~\cite{misra2016cross}\\ Sluice Networks~\cite{ruder2019latent}\\NDDR-CNN~\cite{gao2019nddr}\\MTAN~\cite{liu2019end}\\Branched MTL~\cite{lu2017fully,vandenhende2019branched,bruggemann2020automated,guo2020learning},fill=level4]]
[Decoder-Focused\\ \textbf{(Sec.~\ref{subsec: decoder_focused_architectures})},fill=level3
[PAD-Net~\cite{xu2018pad}\\PAP-Net~\cite{zhang2019pattern}\\JTRL~\cite{zhang2018joint}\\MTI-Net~\cite{vandenhende2020mti}\\PSD~\cite{zhou2020pattern},fill=level4]]
[Other\\\textbf{(Sec.~\ref{subsec: architectures_other})},fill=level3
[ASTMT~\cite{maninis2019attentive},fill=level4]
]
]
[{Optimization Strategy \\ Methods \textbf{(Sec.~\ref{sec: optimization})}},fill=level2
[{Task Balancing\\ \textbf{(Sec.~\ref{subsec: task_balancing})}},fill=level3
[Fixed\\Uncertainty~\cite{kendall2018multi}\\GradNorm~\cite{chen2018gradnorm}\\DWA~\cite{liu2019end}\\DTP~\cite{guo2018dynamic}\\Multi-Objective Optim.~\cite{sener2018multi,lin2019pareto},fill=level4]
]
[{Other\\ \textbf{(Sec.~\ref{subsec: task_balancing_other})}},fill=level3
[Adversarial~\cite{liu2017adversarial,maninis2019attentive,sinha2018gradient}\\Modulation~\cite{zhao2018modulation}\\Heuristics~\cite{sanh2019hierarchical,raffel2019exploring}\\Gradient Sign Dropout~\cite{chen2020just},fill=level4]
]
]
]
\end{forest}
\caption{A taxonomy of deep learning approaches for jointly solving multiple dense prediction tasks.}
\label{fig: taxonomy}
\end{figure*}
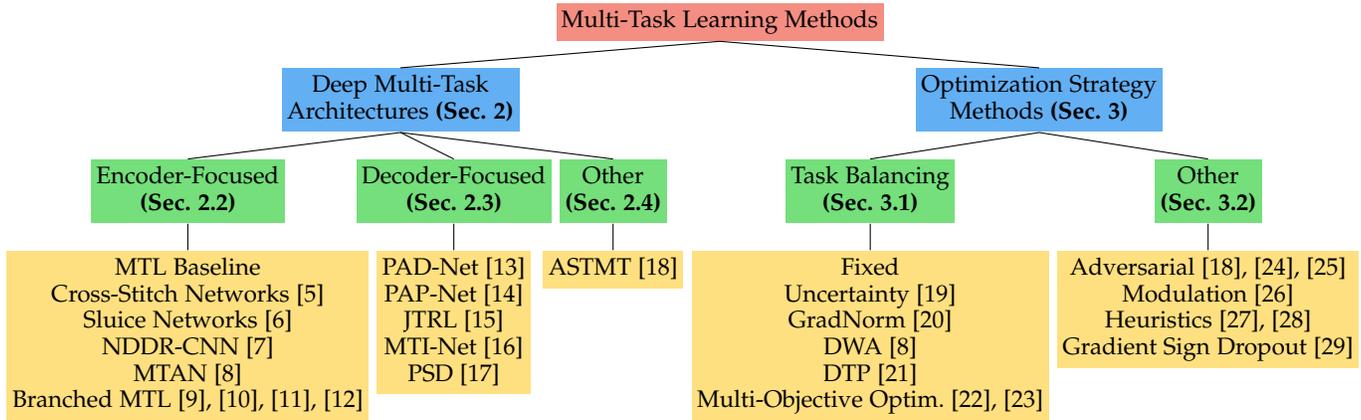

\noindent\textbf{Multi-Task Learning (MTL)}~\cite{caruana1997multitask} aims to improve such generalization by leveraging domain-specific information contained in the training signals of related tasks. In the deep learning era, MTL translates to designing networks capable of learning shared representations from multi-task supervisory signals. Compared to the single-task case, where each individual task is solved separately by its own network, such multi-task networks bring several advantages to the table. First, due to their inherent layer sharing, the resulting memory footprint is substantially reduced. Second, as they explicitly avoid to repeatedly calculate the features in the shared layers, once for every task, they show increased inference speeds. Most importantly, they have the potential for improved performance if the associated tasks share complementary information, or act as a regularizer for one another.

\noindent\textbf{Scope.} 
In this survey, we study deep learning approaches for MTL in computer vision. We refer the interested reader to~\cite{zhang2017survey} for an overview of MTL in other application domains, such as natural language processing~\cite{wang2018glue}, speech recognition~\cite{deng2013new}, bioinformatics~\cite{widmer2010leveraging}, etc. Most importantly, we emphasize on solving multiple \emph{pixel-level} or \emph{dense prediction tasks}, rather than multiple image-level classification tasks, a case that has been mostly under-explored in MTL. Tackling multiple dense prediction tasks differs in several aspects from solving multiple classification tasks. First, as jointly learning multiple dense prediction tasks is governed by the use of different loss functions, unlike classification tasks that mostly use cross-entropy losses, additional consideration is required to avoid a scenario where some tasks overwhelm the others during training. Second, opposed to image-level classification tasks, dense prediction tasks can not be directly predicted from a shared global image representation~\cite{bell2020groknet}, which renders the network design more difficult. Third, pixel-level tasks in scene understanding often have similar characteristics~\cite{zhang2019pattern}, and these similarities can potentially be used to boost the performance under a MTL setup. A popular example is semantic segmentation and depth estimation~\cite{xu2018pad}.

\noindent\textbf{Motivation.} The abundant literature on MTL is rather fragmented. For example, we identify two main groups of works on deep multi-task architectures in Section~\ref{sec: deep_multi_task_architectures} that have been considered largely independent from each other. Moreover, there is limited agreement on the used evaluation metrics and benchmarks. This paper aims to provide a more unified view on the topic. Additionally, we provide a comprehensive experimental study where different groups of works are evaluated in an apples-to-apples comparison. 

\noindent\textbf{Related work.} MTL has been the subject of several surveys~\cite{caruana1997multitask,ruder2017overview,zhang2017survey,gong2019comparison}. In~\cite{caruana1997multitask}, Caruana showed that MTL can be beneficial as it allows for the acquisition of inductive bias through the inclusion of related additional tasks into the training pipeline. The author showcased the use of MTL in artificial neural networks, decision trees and k-nearest neighbors methods, but this study is placed in the very early days of neural networks, rendering it outdated in the deep learning era. Ruder~\cite{ruder2017overview} gave an overview of recent MTL techniques (e.g.~\cite{misra2016cross,lu2017fully,kendall2018multi,ruder2019latent}) applied in deep neural networks. In the same vein, Zhang and Yang~\cite{zhang2017survey} provided a survey that includes feature learning, low-rank, task clustering, task relation learning, and decomposition approaches for MTL. Yet, both works are literature review studies without an empirical evaluation or comparison of the presented techniques. Finally, Gong et al.~\cite{gong2019comparison} benchmarked several optimization techniques (e.g.~\cite{kendall2018multi,liu2019end}) across three MTL datasets. Still, the scope of this study is rather limited, and explicitly focuses on the optimization aspect. Most importantly, all prior studies provide a general overview on MTL without giving specific attention to dense prediction tasks that are of utmost importance in computer vision. 

\noindent\textbf{Paper overview.}
In the following sections, we provide a well-rounded view on state-of-the-art MTL techniques that fall within the defined scope. Section~\ref{sec: deep_multi_task_architectures} considers different deep multi-task architectures, categorizing them into two main groups: encoder- and decoder-focused approaches. Section~\ref{sec: optimization} surveys various optimization techniques for balancing the influence of the tasks when updating the network's weights. We consider the majority of task balancing, adversarial and modulation techniques. In Section~\ref{sec: experiments}, we provide an extensive experimental evaluation across different datasets both within the scope of each group of methods (e.g. encoder-focused approaches) as well as across groups of methods (e.g. encoder- vs decoder-focused approaches). Section~\ref{sec: related_domains} discusses the relations of MTL with other fields. Section~\ref{sec: conclusion} concludes the paper.

Figure~\ref{fig: taxonomy} shows a structured overview of the paper. Our code is made publicly available to ease the adoption of the reviewed MTL techniques: \url{https://github.com/SimonVandenhende/Multi-Task-Learning-PyTorch}.

\begin{figure*}
\begin{minipage}[b]{.48\textwidth}
\centering
\begin{subfigure}[t]{.438\linewidth}
  \centering
  \includegraphics[width=\textwidth]{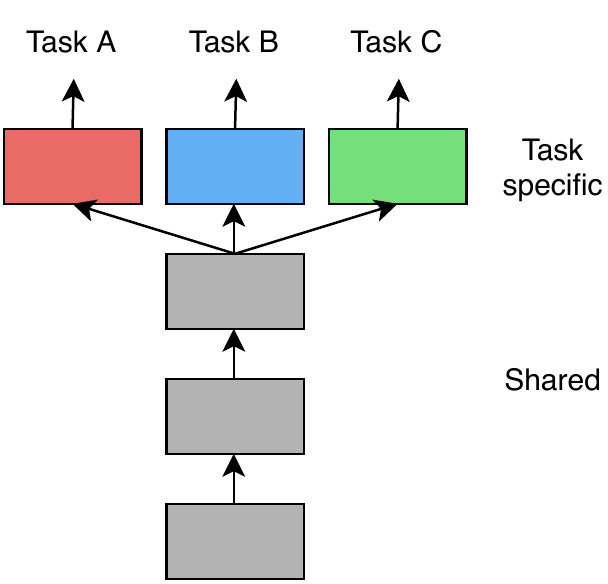}
  \caption{Hard parameter sharing}
  \label{fig: hard_parameter_sharing}
\end{subfigure}%
\hspace*{\fill}
\begin{subfigure}[t]{.523\linewidth}
  \centering
  \includegraphics[width=\textwidth]{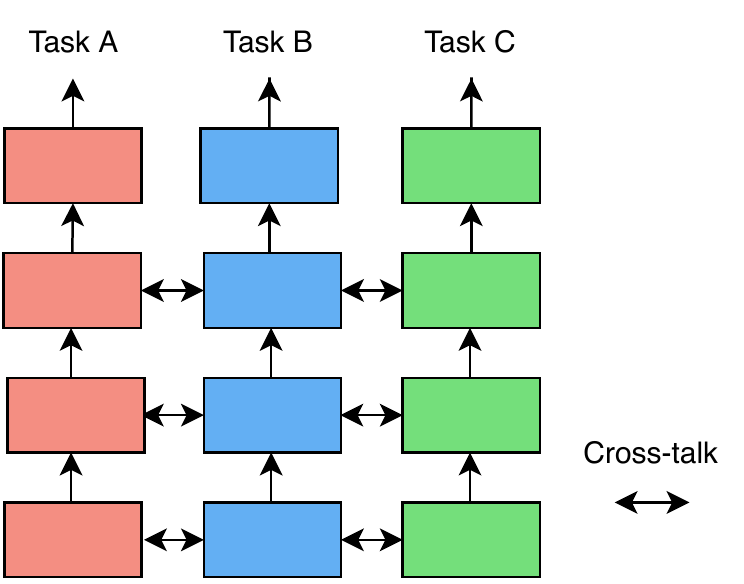}
  \caption{Soft parameter sharing}
  \label{fig: soft_parameter_sharing}
\end{subfigure}
\caption{Historically multi-task learning using deep neural networks has been subdivided into soft- and hard-parameter sharing schemes.}
\label{fig: multi_task_architectures}
\end{minipage}
\hspace*{\fill}
\begin{minipage}[b]{.48\textwidth}
\centering
\begin{subfigure}[t]{.438\linewidth}
  \centering
  \includegraphics[width=\textwidth]{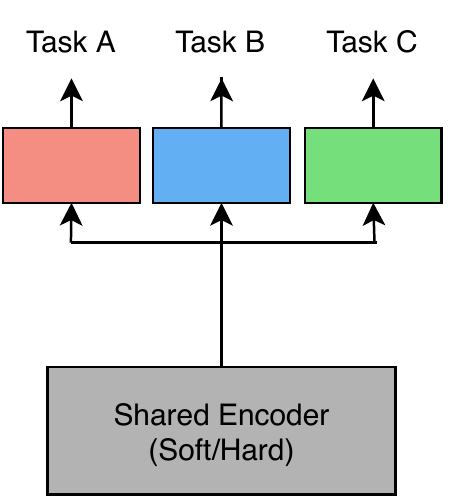}
  \caption{Encoder-focused model}
  \label{fig: encoder_focused_model}
\end{subfigure}%
\hspace*{\fill}
\begin{subfigure}[t]{.438\linewidth}
  \centering
  \includegraphics[width=\textwidth]{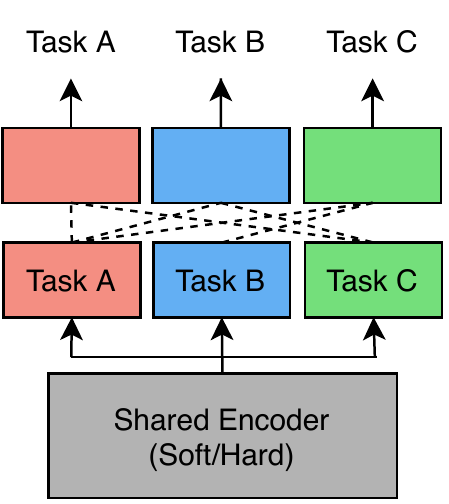}
  \caption{Decoder-focused model}
  \label{fig: decoder_focused_model}
\end{subfigure}
\caption{In this work we discriminate between encoder- and decoder-focused models depending on where the task interactions take place.}
\label{fig: multi_task_architectures_new}
\end{minipage}
\end{figure*}

\section{Deep Multi-Task Architectures}
\label{sec: deep_multi_task_architectures}
In this section, we review deep multi-task architectures used in computer vision. First, we give a brief historical overview of MTL approaches, before introducing a novel taxonomy to categorize different methods. Second, we discuss network designs from different groups of works, and analyze their advantages and disadvantages. An experimental comparison is also provided later in Section~\ref{sec: experiments}. Note that, as a detailed presentation of each architecture is beyond the scope of this survey, in each case we refer the reader to the corresponding paper for further details that complement the following descriptions. 

\subsection{Historical Overview and Taxonomy}
\label{subsec: historical}

\subsubsection{Non-Deep Learning Methods}
\label{subsubsec: non_deep}
Before the deep learning era, MTL works tried to model the common information among tasks in the hope that a joint task learning could result in better generalization performance. To achieve this, they placed assumptions on the task parameter space, such as: task parameters should lie close to each other w.r.t. some distance metric~\cite{evgeniou2004regularized,xue2007multi,jacob2009clustered,zhou2011clustered}, share a common probabilistic prior~\cite{bakker2003task,yu2005learning,lee2007learning,daume2009bayesian,kumar2012learning}, or reside in a low dimensional subspace~\cite{argyriou2008convex,liu2012multi,jalali2010dirty} or manifold~\cite{agarwal2010learning}. These assumptions work well when all tasks are related~\cite{evgeniou2004regularized,ando2005framework,argyriou2008convex,rai2010infinite}, but can lead to performance degradation if information sharing happens between unrelated tasks. The latter is a known problem in MTL, referred to as \emph{negative transfer}. To mitigate this problem, some of these works opted to cluster tasks into groups based on prior beliefs about their similarity or relatedness.

\subsubsection{Soft and Hard Parameter Sharing in Deep Learning} 
\label{subsubsec: soft_vs_hard}
In the context of deep learning, MTL is performed by learning shared representations from multi-task supervisory signals. Historically, deep multi-task architectures were classified into hard or soft parameter sharing techniques. In \textit{hard parameter sharing}, the parameter set is divided into shared and task-specific parameters (see Figure~\ref{fig: hard_parameter_sharing}). MTL models using hard parameter sharing typically consist of a shared encoder that branches out into task-specific heads~\cite{neven2017fast,kendall2018multi,chen2018gradnorm,sener2018multi,teichmann2018multinet}. In \textit{soft parameter sharing}, each task is assigned its own set of parameters and a feature sharing mechanism handles the cross-task talk (see Figure~\ref{fig: soft_parameter_sharing}). We summarize representative works for both groups of works below. 

\noindent\textbf{Hard Parameter Sharing.} UberNet~\cite{kokkinos2017ubernet} was the first hard-parameter sharing model to jointly tackle a large number of low-, mid-, and high-level vision tasks. The model featured a multi-head design across different network layers and scales. Still, the most characteristic hard parameter sharing design consists of a shared encoder that branches out into task-specific decoding heads~\cite{neven2017fast,kendall2018multi,chen2018gradnorm,sener2018multi,teichmann2018multinet}. Multilinear relationship networks~\cite{long2017learning} extended this design by placing tensor normal priors on the parameter set of the fully connected layers. In these works the branching points in the network are determined ad hoc, which can lead to suboptimal task groupings. To alleviate this issue, several recent works~\cite{lu2017fully,vandenhende2019branched,bruggemann2020automated,guo2020learning} proposed efficient design procedures that automatically decide where to share or branch within the network. Similarly, stochastic filter groups~\cite{bragman2019stochastic} re-purposed the convolution kernels in each layer to support shared or task-specific behaviour.

\noindent\textbf{Soft Parameter Sharing.} Cross-stitch networks~\cite{misra2016cross} introduced soft-parameter sharing in deep MTL architectures. The model uses a linear combination of the activations in every layer of the task-specific networks as a means for soft feature fusion. Sluice networks~\cite{ruder2019latent} extended this idea by allowing to learn the selective sharing of layers, subspaces and skip connections. NDDR-CNN~\cite{gao2019nddr} also incorporated dimensionality reduction techniques into the feature fusion layers. Differently, MTAN~\cite{liu2019end} used an attention mechanism to share a general feature pool amongst the task-specific networks. A concern with soft parameter sharing approaches is scalability, as the size of the multi-task network tends to grow linearly with the number of tasks.

\subsubsection{Distilling Task Predictions in Deep Learning}
\label{subsubsec: distilling}
All works presented in Section~\ref{subsubsec: soft_vs_hard} follow a common pattern: they \textit{directly} predict all task outputs from the same input in one processing cycle. In contrast, a few recent works first employed a multi-task network to make initial task predictions, and then leveraged features from these initial predictions to further improve each task output -- in a one-off or recursive manner. PAD-Net~\cite{xu2018pad} proposed to distill information from the initial task predictions of other tasks, by means of spatial attention, before adding it as a residual to the task of interest. JTRL~\cite{zhang2018joint} opted for sequentially predicting each task, with the intention to utilize information from the past predictions of one task to refine the features of another task at each iteration. PAP-Net~\cite{zhang2019pattern} extended upon this idea, and used a recursive procedure to propagate similar cross-task and task-specific patterns found in the initial task predictions. To do so, they operated on the affinity matrices of the initial predictions, and not on the features themselves, as was the case before~\cite{xu2018pad,zhang2018joint}. Zhou et al.~\cite{zhou2020pattern} refined the use of pixel affinities to distill the information by separating inter- and intra-task patterns from each other. MTI-Net~\cite{vandenhende2020mti} adopted a multi-scale multi-modal distillation procedure to explicitly model the unique task interactions that happen at each individual scale. 

\subsubsection{A New Taxonomy of MTL Approaches}
As explained in Section~\ref{subsubsec: soft_vs_hard}, multi-task networks have historically been classified into soft or hard parameter sharing techniques. However, several recent works took inspiration from both groups of works to jointly solve multiple pixel-level tasks. As a consequence, it is debatable whether the soft vs hard parameter sharing paradigm should still be used as the main framework for classifying MTL architectures. In this survey, we propose an alternative taxonomy that discriminates between different architectures on the basis of where the \emph{task interactions} take place, i.e. locations in the network where information or features are exchanged or shared between tasks. The impetus for this framework was given in Section~\ref{subsubsec: distilling}. Based on the proposed criterion, we distinguish between two types of models: \emph{encoder-focused} and \emph{decoder-focused} architectures. The encoder-focused architectures (see Figure~\ref{fig: encoder_focused_model}) only share information in the encoder, using either hard- or soft-parameter sharing, before decoding each task with an independent task-specific head. Differently, the decoder-focused architectures (see Figure~\ref{fig: decoder_focused_model}) also exchange information during the decoding stage. Figure~\ref{fig: taxonomy} gives an overview of the proposed taxonomy, listing representative works in each case.

\subsection{Encoder-focused Architectures}
\label{subsec: encoder_focused_architectures}
Encoder-focused architectures (see Figure~\ref{fig: encoder_focused_model}) share the task features in the encoding stage, before they process them with a set of independent task-specific heads. A number of works~\cite{neven2017fast,kendall2018multi,chen2018gradnorm,sener2018multi,teichmann2018multinet} followed an ad hoc strategy by sharing an off-the-shelf backbone network in combination with small task-specific heads (see Figure~\ref{fig: hard_parameter_sharing}). This model relies on the encoder (i.e. backbone network) to learn a generic representation of the scene. The features from the encoder are then used by the task-specific heads to get the predictions for every task. While this simple model shares the full encoder amongst all tasks, recent works have considered \textit{where} and \textit{how} the feature sharing should happen in the encoder. We discuss such sharing strategies in the following sections.

\begin{figure}
\begin{minipage}[t]{0.50\linewidth}
\centering
\includegraphics[width=\textwidth]{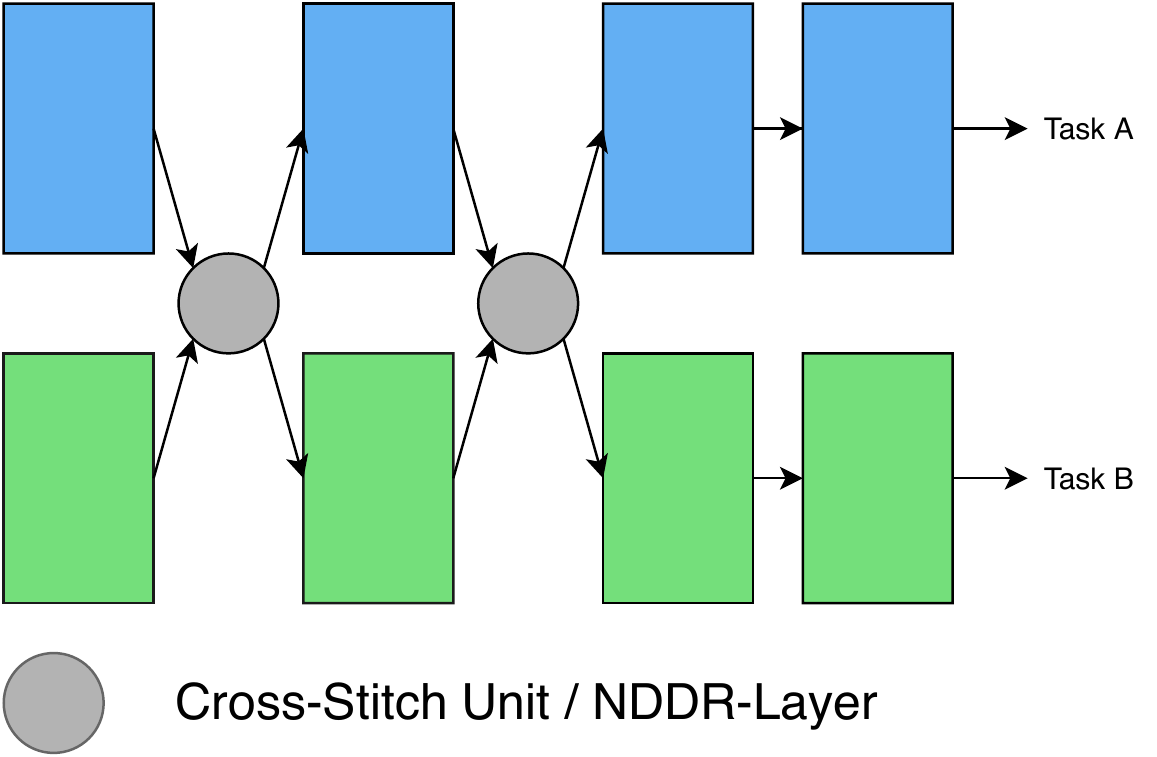}
\caption{The architecture of cross-stitch networks~\cite{misra2016cross} and NDDR-CNNs~\cite{gao2019nddr}. The activations from all single-task networks are fused across several encoding layers. Different feature fusion mechanisms were used in each case.}
\label{fig: crossstitch_nddr}
\end{minipage}
\hspace*{\fill}
\begin{minipage}[t]{0.47\linewidth}
\centering
\includegraphics[width=\textwidth]{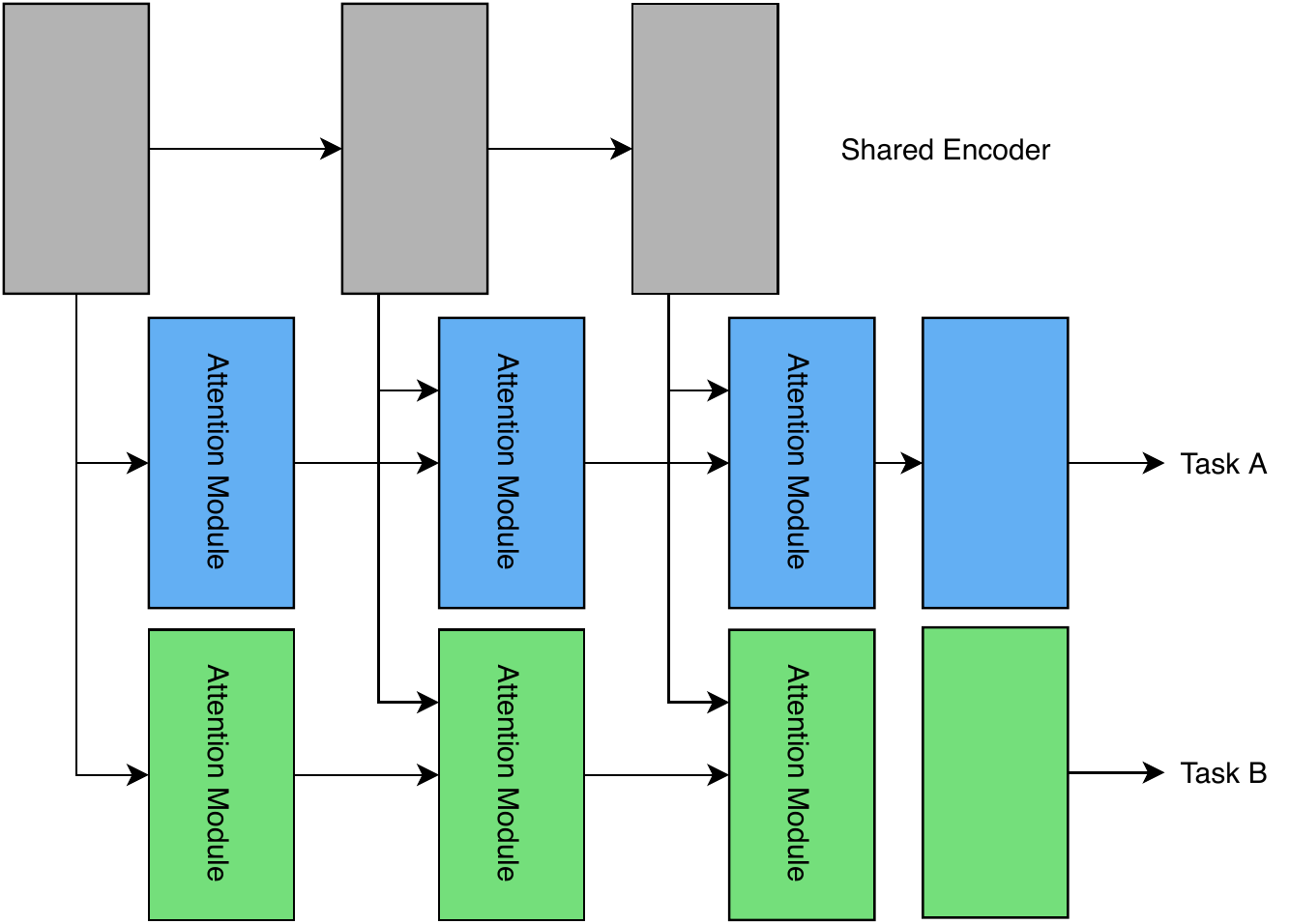}
\caption{The architecture of MTAN~\cite{liu2019end}. Task-specific attention modules select and refine features from several layers of a shared encoder.}
\label{fig: mtan}
\end{minipage}
\end{figure}

\subsubsection{Cross-Stitch Networks}
\label{subsubsec: cross_stitch_nets}

\textbf{Cross-stitch networks}~\cite{misra2016cross} shared the activations amongst all single-task networks in the encoder. Assume we are given two activation maps $x_A, x_B$ at a particular layer, that belong to tasks $A$ and $B$ respectively. A learnable linear combination of these activation maps is applied, before feeding the transformed result $\tilde{x}_A, \tilde{x}_B$ to the next layer in the single-task networks. The transformation is parameterized by learnable weights $\alpha$, and can be expressed as
\begin{equation}
\begin{bmatrix}
\tilde{x}_A \\
\tilde{x}_B
\end{bmatrix}
= 
\begin{bmatrix}
\alpha_{AA} & \alpha_{AB} \\
\alpha_{BA} & \alpha_{BB} \\
\end{bmatrix}
\begin{bmatrix}
x_A \\
x_B
\end{bmatrix}.
\end{equation}
As illustrated in Figure~\ref{fig: crossstitch_nddr}, this procedure is repeated at multiple locations in the encoder. By learning the weights $\alpha$, the network can decide the degree to which the features are shared between tasks. In practice, we are required to pre-train the single-task networks, before stitching them together, in order to maximize the performance. A disadvantage of cross-stitch networks is that the size of the network increases linearly with the number of tasks. Furthermore, it is not clear where the cross-stitch units should be inserted in order to maximize their effectiveness. \textbf{Sluice networks}~\cite{ruder2019latent} extended this work by also supporting the selective sharing of subspaces and skip connections. 

\subsubsection{Neural Discriminative Dimensionality Reduction}
\label{subsubsec: nddr_cnn}

\textbf{Neural Discriminative Dimensionality Reduction CNNs} (NDDR-CNNs)~\cite{gao2019nddr} used a similar architecture with cross-stitch networks (see Figure~\ref{fig: crossstitch_nddr}). However, instead of utilizing a linear combination to fuse the activations from all single-task networks, a dimensionality reduction mechanism is employed. First, features with the same spatial resolution in the single-task networks are concatenated channel-wise. Second, the number of channels is reduced by processing the features with a 1 by 1 convolutional layer, before feeding the result to the next layer. The convolutional layer allows to fuse activations across all channels. Differently, cross-stitch networks only allow to fuse activations from channels that share the same index. The NDDR-CNN behaves as a cross-stitch network when the non-diagonal elements in the weight matrix of the convolutional layer are zero.

Due to their similarity with cross-stitch networks, NDDR-CNNs are prone to the same problems. First, there is a scalability concern when dealing with a large number of tasks. Second, NDDR-CNNs involve additional design choices, since we need to decide where to include the NDDR layers. Finally, both cross-stitch networks and NDDR-CNNs only allow to use limited local information (i.e. small receptive field) when fusing the activations from the different single-task networks. We hypothesize that this is suboptimal because the use of sufficient context is very important during encoding -- as already shown for the tasks of image classification~\cite{he2015spatial} and semantic segmentation~\cite{chen2017deeplab,zhao2017pyramid,yuan2018ocnet}. This is backed up by certain decoder-focused architectures in Section~\ref{subsec: decoder_focused_architectures} that overcome the limited receptive field by predicting the tasks at multiple scales and by sharing the features repeatedly at every scale. 

\subsubsection{Multi-Task Attention Networks}
\label{subsubsec: mtan}

\textbf{Multi-Task Attention Networks} (MTAN)~\cite{liu2019end} used a shared backbone network in conjunction with task-specific attention modules in the encoder (see Figure~\ref{fig: mtan}). The shared backbone extracts a general pool of features. Then, each task-specific attention module selects features from the general pool by applying a soft attention mask. The attention mechanism is implemented using regular convolutional layers and a sigmoid non-linearity. Since the attention modules are small compared to the backbone network, the MTAN model does not suffer as severely from the scalability issues that are typically associated with cross-stitch networks and NDDR-CNNs. However, similar to the fusion mechanism in the latter works, the MTAN model can only use limited local information to produce the attention mask.  

\subsubsection{Branched Multi-Task Learning Networks}
\label{subsubsec: branched_mtl}

The models presented in Sections~\ref{subsubsec: cross_stitch_nets}-\ref{subsubsec: mtan} softly shared the features amongst tasks during the encoding stage. Differently, branched multi-task networks followed a hard-parameter sharing scheme. Before presenting these methods, consider the following observation: deep neural networks tend to learn hierarchical image representations~\cite{yosinski2014transferable}. The early layers tend to focus on more general low-level image features, such as edges, corners, etc., while the deeper layers tend to extract high-level information that is more task-specific. Motivated by this observation, branched MTL networks opted to learn similar hierarchical encoding structures~\cite{lu2017fully,vandenhende2019branched,bruggemann2020automated,guo2020learning}. These ramified networks typically start with a number of shared layers, after which different (groups of) tasks branch out into their own sequence of layers. In doing so, the different branches gradually become more task-specific as we move to the deeper layers. This behaviour aligns well with the hierarchical representations learned by deep neural nets. However, as the number of possible network configurations is combinatorially large, deciding what layers to share and where to branch out becomes cumbersome. Several works have tried to automate the procedure of hierarchically clustering the tasks to form branched MTL networks given a specific computational budget (e.g. number of parameters, FLOPS). We provide a summary of existing works below. 

\textbf{Fully-Adaptive Feature Sharing (FAFS)}~\cite{lu2017fully} starts from a network where tasks initially share all layers, and dynamically grows the model in a greedy layer-by-layer fashion during training. The task groupings are optimized to separate dissimilar tasks from each other, while minimizing network complexity. The task relatedness is based on the probability of concurrently 'simple' or 'difficult' examples across tasks. This strategy assumes that it is preferable to solve two tasks in an isolated manner (i.e. different branches) when the majority of examples are 'simple' for one task, but 'difficult' for the other. 

Similar to FAFS, Vandenhende et al.~\cite{vandenhende2019branched} rely on pre-computed task relatedness scores to decide the grouping of tasks. In contrast to FAFS, they measure the task relatedness based on feature affinity scores, rather than sample difficulty. The main assumption is that two tasks are strongly related, if their single-task models rely on a similar set of features. An efficient method~\cite{dwivedi2019representation} is used to quantify this property. An advantage over FAFS is that the task groupings can be determined offline for the whole network, and not online in a greedy layer-by-layer fashion~\cite{vandenhende2019branched}. This strategy promotes task groupings that are optimal in a global, rather than local, sense. Yet, a disadvantage is that the calculation of the task affinity scores requires a set of single-task networks to be pretrained first.  

Different from the previous works, \textbf{Branched Multi-Task Architecture Search} (BMTAS)~\cite{bruggemann2020automated} and \textbf{Learning To Branch} (LTB)~\cite{guo2020learning} have directly optimized the network topology without relying on pre-computed task relatedness scores. More specifically, they rely on a tree-structured network design space where the branching points are casted as a Gumbel softmax operation. This strategy has the advantage over~\cite{lu2017fully,vandenhende2019branched} that the task groupings can be directly optimized end-to-end for the tasks under consideration. Moreover, both methods can easily be applied to any set of tasks, including both image classification and per-pixel prediction tasks. Similar to~\cite{lu2017fully,vandenhende2019branched}, a compact network topology can be obtained by including a resource-aware loss term. In this case, the computational budget is jointly optimized with the multi-task learning objective in an end-to-end fashion.

\subsection{Decoder-Focused Architectures}
\label{subsec: decoder_focused_architectures}

The encoder-focused architectures in Section~\ref{subsec: encoder_focused_architectures} follow a common pattern: they \textit{directly} predict all task outputs from the same input in one processing cycle (i.e. all predictions are generated once, in parallel or sequentially, and are not refined afterwards). By doing so, they fail to capture commonalities and differences among tasks, that are likely fruitful for one another (e.g. depth discontinuities are usually aligned with semantic edges). Arguably, this might be the reason for the moderate only performance improvements achieved by the encoder-focused approaches to MTL (see Section~\ref{subsubsec: experiments_encoder}). To alleviate this issue, a few recent works first employed a multi-task network to make initial task predictions, and then leveraged features from these initial predictions in order to further improve each task output -- in an one-off or recursive manner. As these MTL approaches also share or exchange information during the decoding stage, we refer to them as decoder-focused architectures (see Figure~\ref{fig: decoder_focused_model}). 

\subsubsection{PAD-Net}
\begin{figure}[t]
    \centering
    \includegraphics[width=.9\linewidth]{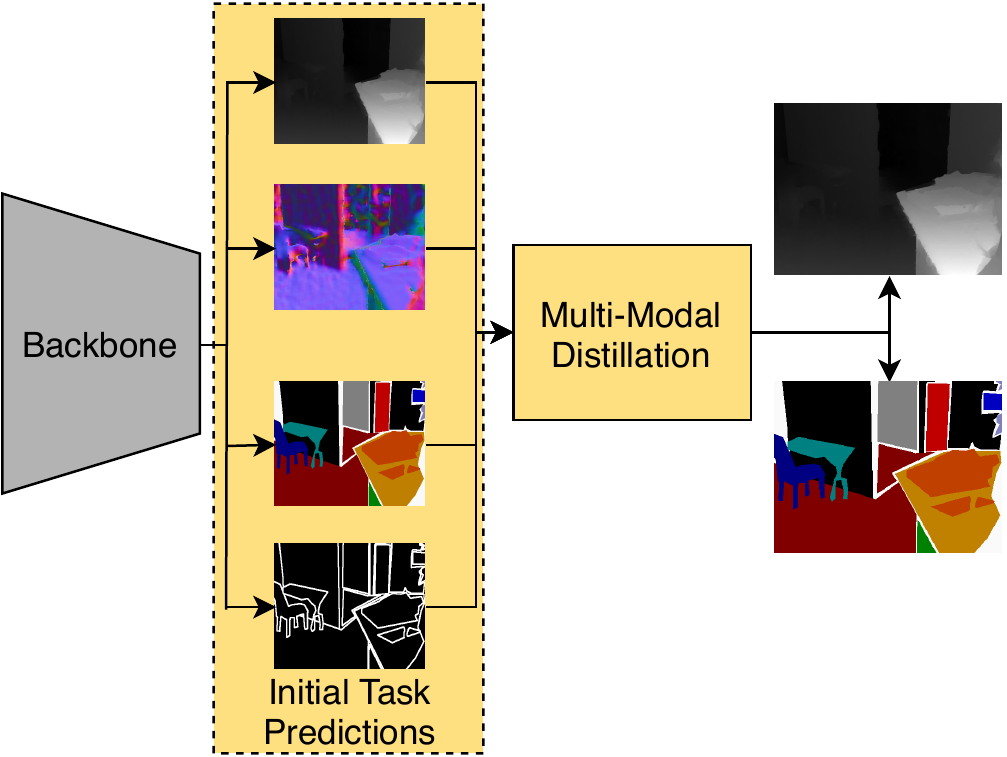}
    \caption{The architecture in PAD-Net~\cite{xu2018pad}. Features extracted by a backbone network are passed to task-specific heads to make initial task predictions. The task features from the different heads are then combined through a distillation unit to make the final predictions. Note that, auxiliary tasks can be used in this framework, i.e. tasks for which only the initial predictions are generated, but not the final ones.}
    \label{fig: padnet}
\end{figure}

\textbf{PAD-Net}~\cite{xu2018pad} was one of the first decoder-focused architectures. The model itself is visualized in Figure~\ref{fig: padnet}. As can be seen, the input image is first processed by an off-the-shelf backbone network. The backbone features are further processed by a set of task-specific heads that produce an initial prediction for every task. These initial task predictions add deep supervision to the network, but they can also be used to exchange information between tasks, as will be explained next. The \textit{task features} in the last layer of the task-specific heads contain a per-task feature representation of the scene. PAD-Net proposed to re-combine them via a multi-modal distillation unit, whose role is to extract cross-task information, before producing the final task predictions. 

PAD-Net performs the multi-modal distillation by means of a spatial attention mechanism. Particularly, the output features $F_{k}^o$ for task $k$ are calculated as
\begin{equation}
\label{eq: spatial_attention}
F_{k}^{o} = F_{k}^i + \sum_{l \neq k} \sigma \left(W_{k,l} F_{l}^{i} \right) \odot F_l^i,
\end{equation}
where $\sigma \left(W_{k,l} F_{l}^{i} \right)$ returns a spatial attention mask that is applied to the initial task features $F_{l}^i$ from task $l$. The attention mask itself is found by applying a convolutional layer $W_{k,l}$ to extract local information from the initial task features. Equation~\ref{eq: spatial_attention} assumes that the task interactions are location dependent, i.e. tasks are not in a constant relationship across the entire image. This can be understood from a simple example. Consider two dense prediction tasks, e.g. monocular depth prediction and semantic segmentation. Depth discontinuities and semantic boundaries often coincide. However, when we segment a flat object, e.g. a magazine, from a flat surface, e.g. a table, we will still find a semantic boundary where the depth map is rather continuous. In this particular case, the depth features provide no additional information for the localization of the semantic boundaries. The use of spatial attention explicitly allows the network to select information from other tasks at locations where its useful. 

The encoder-focused approaches in Section~\ref{subsec: encoder_focused_architectures} shared features amongst tasks using the intermediate representations in the encoder. Differently, PAD-Net models the task interactions by applying a spatial attention layer to the features in the task-specific heads. In contrast to the intermediate feature representations in the encoder, the task features used by PAD-Net are already disentangled according to the output task. We hypothesize that this makes it easier for other tasks to distill the relevant information. This multi-step decoding strategy from PAD-Net is applied and refined in other decoder-focused approaches.  

\subsubsection{Pattern-Affinitive Propagation Networks}

\begin{figure}
    \centering
    \includegraphics[width=\linewidth]{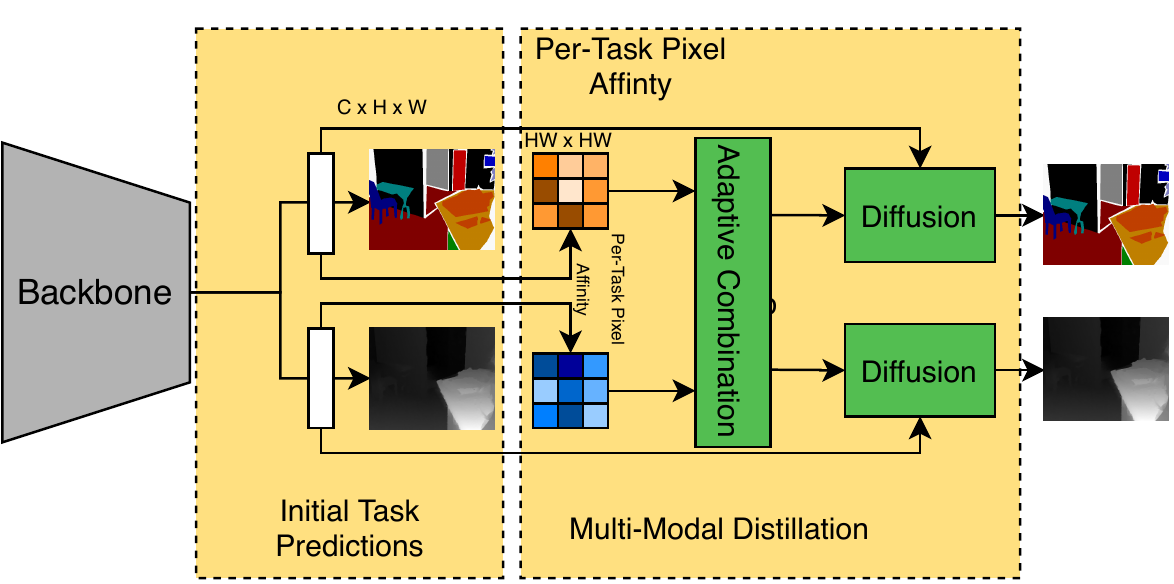}
    \caption{The architecture in PAP-Net~\cite{zhang2019pattern}. Features extracted by a backbone network are passed to task-specific heads to make initial task predictions. The task features from the different heads are used to calculate a per-task pixel affinity matrix. The affinity matrices are adaptively combined and diffused back into the task features space to spread the cross-task correlation information across the image. The refined features are used to make the final task predictions.}
    \label{fig: papnet}
\end{figure}

\textbf{Pattern-Affinitive Propagation Networks} (PAP-Net)~\cite{zhang2019pattern} used an architecture similar to PAD-Net (see Figure~\ref{fig: papnet}), but the multi-modal distillation in this work is performed in a different manner. The authors argue that directly working on the task features space via the spatial attention mechanism, as done in PAD-Net, might be a suboptimal choice. As the optimization still happens at a different space, i.e. the task label space, there is no guarantee that the model will learn the desired task relationships. Instead, they statistically observed that pixel affinities tend to align well with common local structures on the task label space. Motivated by this observation, they proposed to leverage pixel affinities in order to perform multi-modal distillation.

To achieve this, the backbone features are first processed by a set of task-specific heads to get an initial prediction for every task. Second, a per-task pixel affinity matrix $M_{T_j}$ is calculated by estimating pixel-wise correlations upon the task features coming from each head. Third, a cross-task information matrix $\hat{M}_{T_j}$ for every task $T_j$ is learned by adaptively combining the affinity matrices $M_{T_{i}}$ for tasks $T_i$ with learnable weights $\alpha_i^{T_j}$ 
\begin{equation}
\hat{M}_{T_j} = \sum_{T_i} \alpha_i^{T_j} \cdot M_{T_i}.
\end{equation}
Finally, the task features coming from each head $j$ are refined using the cross-task information matrix $\hat{M}_{T_j}$. In particular, the cross-task information matrix is diffused into the task features space to spread the correlation information across the image. This effectively weakens or strengthens the pixel correlations for task $T_j$, based on the pixel affinities from other tasks $T_i$. The refined features are used to make the final predictions for every task. 

All previously discussed methods only use limited local information when fusing features from different tasks. For example, cross-stitch networks and NDDR-CNNs combine the features in a channel-wise fashion, while PAD-Net only uses the information from within a 3 by 3 pixels window to construct the spatial attention mask. Differently, PAP-Net also models the non-local relationships through pixel affinities measured across the entire image. Zhou et al.~\cite{zhou2020pattern} extended this idea to specifically mine and propagate both inter- and intra- task patterns.

\subsubsection{Joint Task-Recursive Learning}

\begin{figure}
    \centering
    \includegraphics[width=\linewidth]{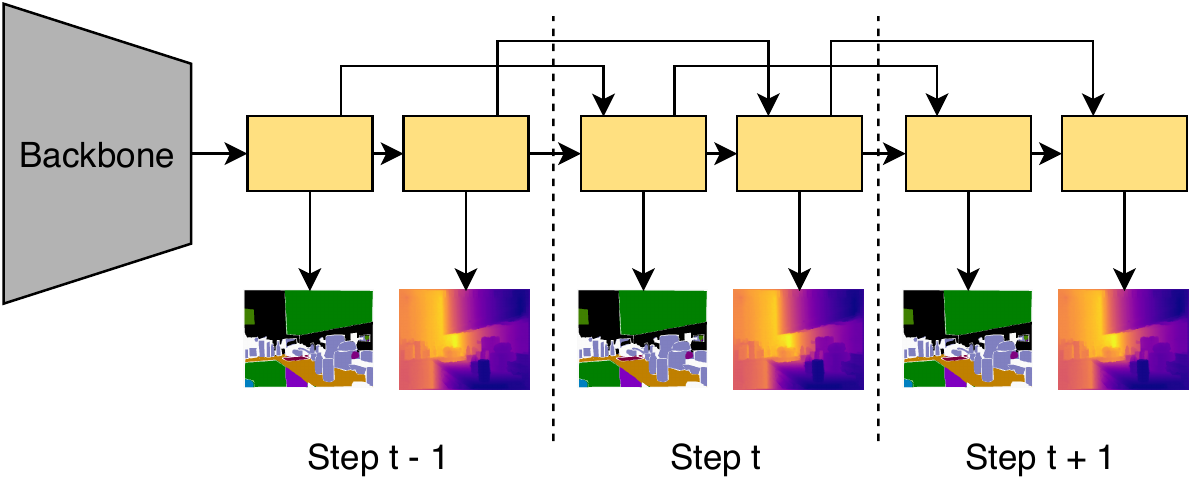}
    \caption{The architecture in Joint Task-Recursive Learning~\cite{zhang2018joint}. The features of two tasks are progressively refined in an intertwined manner based on past states.}
    \label{fig: jtrl}
\end{figure}

\textbf{Joint Task-Recursive Learning} (JTRL)~\cite{zhang2018joint} recursively predicts two tasks at increasingly higher scales in order to gradually refine the results based on past states. The architecture is illustrated in Figure~\ref{fig: jtrl}. Similarly to PAD-Net and PAP-Net, a multi-modal distillation mechanism is used to combine information from earlier task predictions, through which later predictions are refined. Differently, the JTRL model predicts two tasks sequentially, rather than in parallel, and in an intertwined manner. The main disadvantage of this approach is that it is not straightforward, or even possible, to extent this model to more than two tasks given the intertwined manner at which task predictions are refined. 

\subsubsection{Multi-Scale Task Interaction Networks}

In the decoder-focused architectures presented so far, the multi-modal distillation was performed at a fixed scale, i.e. the features of the backbone's last layer. This rests on the assumption that all relevant task interactions can solely be modeled through a single filter operation with a specific receptive field. However, \textbf{Multi-Scale Task Interaction Networks} (MTI-Net)~\cite{vandenhende2020mti} showed that this is a rather strict assumption. In fact, tasks can influence each other differently at different receptive fields.

\begin{figure}[t]
    \centering
    \includegraphics[width=\linewidth]{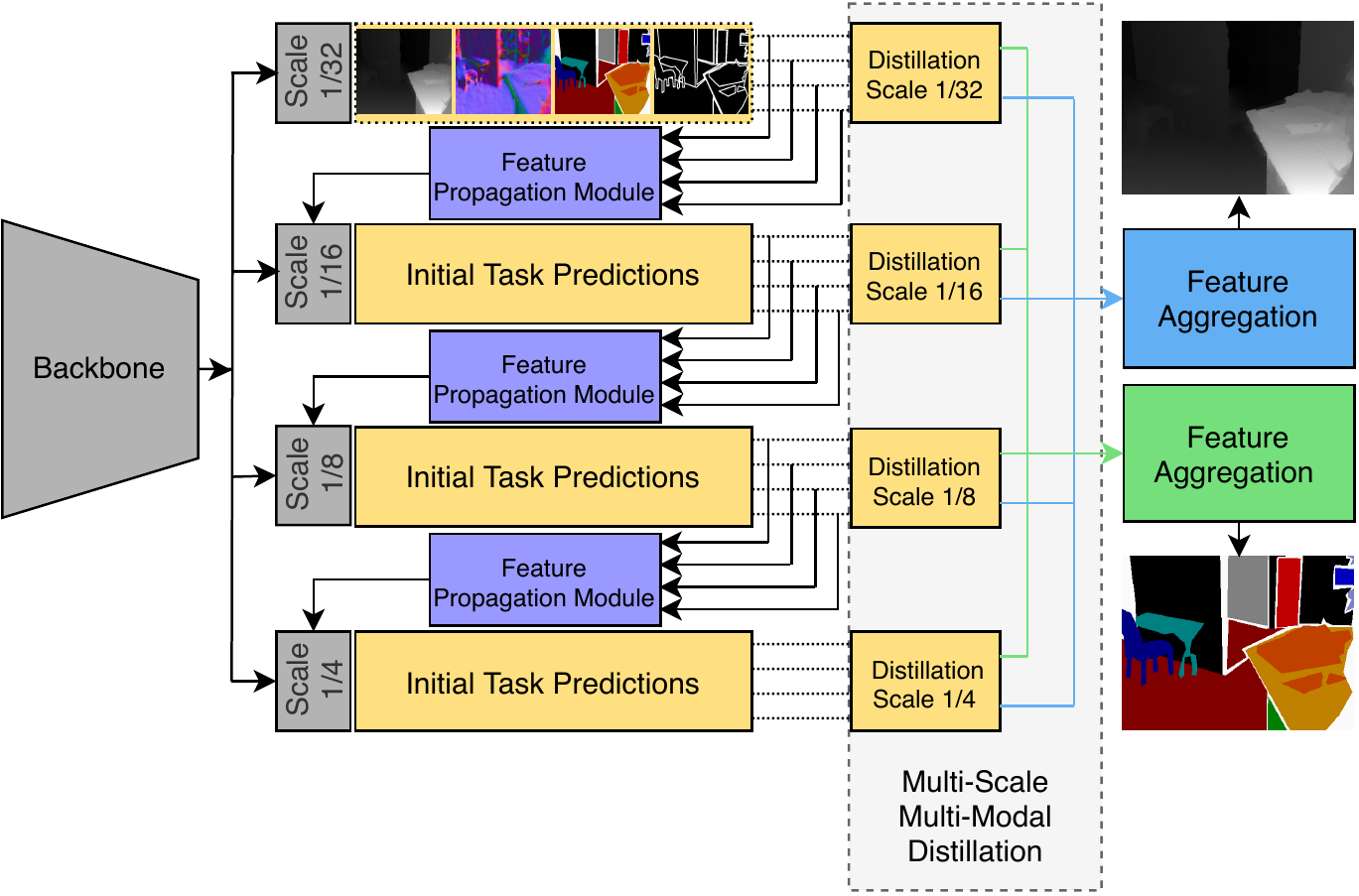}
    \caption{The architecture in Multi-Scale Task Interaction Networks~\cite{vandenhende2020mti}. Starting from a backbone that extracts multi-scale features, initial task predictions are made at each scale. These task features are then distilled separately at every scale, allowing the model to capture unique task interactions at multiple scales, i.e. receptive fields. After distillation, the distilled task features from all scales are aggregated to make the final task predictions. To boost performance, a feature propagation module is included to pass information from lower resolution task features to higher ones. }
    \label{fig: mti_net}
\end{figure}

To account for this restriction, MTI-Net explicitly took into account task interactions at multiple scales. Its architecture is illustrated in Figure~\ref{fig: mti_net}. First, an off-the-shelf backbone network extracts a multi-scale feature representation from the input image. From the multi-scale feature representation an initial prediction for every task is made at each scale. The task predictions at a particular scale are found by applying a task-specific head to the backbone features extracted at that scale. Similarly to PAD-Net, the features in the last layer of the task-specific heads are combined and refined to make the final predictions. Differently, in MTI-Net the per-task feature representations can be distilled at each scale separately. This allows to have multiple task interactions, each modeled within a specific receptive field. The distilled multi-scale features are upsampled to the highest scale and concatenated, resulting in a final feature representation for every task. The final task predictions are found by decoding these final feature representations in a task-specific manner again. The performance was further improved by also propagating information from the lower-resolution task features to the higher-resolution ones using a Feature Propagation Module. 

The experimental evaluation in~\cite{vandenhende2020mti} shows that distilling task information at multiple scales increases the multi-tasking performance compared to PAD-Net where such information is only distilled at a single scale. Furthermore, since MTI-Net distills the features at multiple scales, i.e. using different pixel dilations, it overcomes the issue of using only limited local information to fuse the features, which was already shown to be beneficial in PAP-Net. 

\subsection{Other Approaches}
\label{subsec: architectures_other}

A number of approaches that fall outside the aforementioned categories have been proposed in the literature. For example, multilinear relationship networks~\cite{long2017learning} used tensor normal priors to the parameter set of the task-specific heads to allow interactions in the decoding stage. Different from the standard parallel ordering scheme, where layers are aligned and shared (e.g.~\cite{misra2016cross,gao2019nddr}), soft layer ordering~\cite{meyerson2018beyond} proposed a flexible sharing scheme across tasks and network depths. Yang et al.~\cite{yang2016deep} generalized matrix factorisation approaches to MTL in order to learn cross-task sharing structures in every layer of the network. Routing networks~\cite{rosenbaum2018routing} proposed a principled approach to determine the connectivity of a network's function blocks through routing. Piggyback~\cite{mallya2018piggyback} showed how to adapt a single, fixed neural network to a multi-task network by learning binary masks. Huang et al.~\cite{huang2018gnas} introduced a method rooted in Neural Architecture Search (NAS) for the automated construction of a tree-based multi-attribute learning network. Stochastic filter groups~\cite{bragman2019stochastic} re-purposed the convolution kernels in each layer of the network to support shared or task-specific behaviour. In a similar vein, feature partitioning~\cite{newell2019feature} presented partitioning strategies to assign the convolution kernels in each layer of the network into different tasks. In general, these works have a different scope within MTL, e.g. automate the network architecture design. Moreover, they mostly focus on solving multiple (binary) classification tasks, rather than multiple dense predictions tasks. As a result, they fall outside the scope of this survey, with one notable exception that is discussed next.

\noindent\textbf{Attentive Single-Tasking of Multiple Tasks} (ASTMT)~\cite{maninis2019attentive} proposed to take a 'single-tasking' route for the MTL problem. That is, within a multi-tasking framework they performed separate forward passes, one for each task, that activate shared responses among all tasks, plus some residual responses that are task-specific. Furthermore, to suppress the negative transfer issue they applied adversarial training on the gradients level that enforces them to be statistically indistinguishable across tasks. An advantage of this approach is that shared and task-specific information within the network can be naturally disentangled. On the negative side, however, the tasks can not be predicted altogether, but only one after the other, which significantly increases the inference speed and somehow defies the purpose of MTL. 

\section{Optimization in MTL}
\label{sec: optimization}

In the previous section, we discussed about the construction of network architectures that are able to learn multiple tasks concurrently. Still, a significant challenge in MTL stems from the optimization procedure itself. In particular, we need to carefully balance the joint learning of all tasks to avoid a scenario where one or more tasks have a dominant influence in the network weights. In this section, we discuss several methods that have considered this \textit{task balancing} problem. 

\subsection{Task Balancing Approaches}
\label{subsec: task_balancing}

Without loss of generality, the optimization objective in a MTL problem, assuming task-specific weights $w_i$ and task-specific loss functions $\mathcal{L}_i$, can be formulated as 
\begin{equation}
\label{eq: mtl_loss}
\mathcal{L}_{MTL} = \sum_{i} w_i \cdot \mathcal{L}_i.
\end{equation}
When using stochastic gradient descent to minimize the objective from Equation~\ref{eq: mtl_loss}, which is the standard approach in the deep learning era, the network weights in the shared layers $W_{sh}$ are updated by the following rule
\begin{equation}
\label{eq: update_rule}
W_{sh} = W_{sh} - \gamma \sum_{i} w_i \pdv{\mathcal{L}_i}{W_{sh}}.
\end{equation}
From Equation~\ref{eq: update_rule} we can draw the following conclusions. First, the network weight update can be suboptimal when the task gradients conflict, or dominated by one task when its gradient magnitude is much higher w.r.t. the other tasks. This motivated researchers~\cite{kendall2018multi,guo2018dynamic,liu2019end,chen2018gradnorm} to balance the gradient magnitudes by setting the task-specific weights $w_i$ in the loss. To this end, other works~\cite{sener2018multi,zhao2018modulation,suteu2019regularizing} have also considered the influence of the direction of the task gradients. Second, each task's influence on the network weight update can be controlled, either \textit{indirectly} by adapting the task-specific weights $w_i$ in the loss, or \textit{directly} by operating on the task-specific gradients $\pdv{\mathcal{L}_i}{W_{sh}}$. A number of methods that tried to address these problems are discussed next. 

\subsubsection{Uncertainty Weighting}
\label{subsubsec: uncertainty_weighting}

Kendall et al.~\cite{kendall2018multi} used the \textbf{homoscedastic uncertainty} to balance the single-task losses. The homoscedastic uncertainty or \textit{task-dependent uncertainty} is not an output of the model, but a quantity that remains constant for different input examples of the same task. The optimization procedure is carried out to maximise a Gaussian likelihood objective that accounts for the homoscedastic uncertainty. In particular, they optimize the model weights $W$ and the noise parameters $\sigma_1$, $\sigma_2$ to minimize the following objective 
\begin{equation}
\label{eq: uncertainty_weighting}
\mathcal{L}\left(W,\sigma_1,\sigma_2\right) = \frac{1}{2\sigma^2_1} \mathcal{L}_1 \left(W\right) + \frac{1}{2\sigma^2_2} \mathcal{L}_2 \left(W\right) + \log \sigma_1 \sigma_2.
\end{equation}
The loss functions $\mathcal{L}_1$, $\mathcal{L}_2$ belong to the first and second task respectively. By minimizing the loss $\mathcal{L}$ w.r.t. the noise parameters $\sigma_1$, $\sigma_2$, one can essentially balance the task-specific losses during training. The optimization objective in Equation~\ref{eq: uncertainty_weighting} can easily be extended to account for more than two tasks too. The noise parameters are updated through standard backpropagation during training. 

Note that, increasing the noise parameter $\sigma_i$ reduces the weight for task $i$. Consequently, the effect of task $i$ on the network weight update is smaller when the task's homoscedastic uncertainty is high. This is advantageous when dealing with noisy annotations since the task-specific weights will be lowered automatically for such tasks.  

\subsubsection{Gradient Normalization}
\label{subsubsec: gradient_normalization}

\textbf{Gradient normalization} (GradNorm)~\cite{chen2018gradnorm} proposed to control the training of multi-task networks by stimulating the task-specific gradients to be of similar magnitude. By doing so, the network is encouraged to learn all tasks at an equal pace. Before presenting this approach, we introduce the necessary notations in the following paragraph.

We define the L2 norm of the gradient for the weighted single-task loss $w_i\left(t\right) \cdot L_i\left(t\right)$ at step $t$ w.r.t. the weights $W$, as $G_i^W\left(t\right)$. We additionally define the following quantities, 
\begin{itemize}
\item the mean task gradient $\bar{G}^{W}$ averaged across all task gradients $G_i^{W}$ w.r.t the weights $W$ at step $t$: $\bar{G}^{W} \left(t\right) = E_{task}\left[G_i^W\left(t\right)\right]$;
\item the inverse training rate $\tilde{L}_{i}$ of task $i$ at step $t$: $ \tilde{L}_{i}\left(t\right) = L_i \left(t\right) / L_i\left(0\right)$;
\item the relative inverse training rate of task $i$ at step $t$: $r_i\left(t\right) = \tilde{L}_i\left(t\right) / E_{task}\left[\tilde{L}_i\left(t\right)\right]$.
\end{itemize}

GradNorm aims to balance two properties during the training of a multi-task network. First, balancing the gradient magnitudes $G_i^W$. To achieve this, the mean gradient $\bar{G}^W$ is considered as a common basis from which the relative gradient sizes across tasks can be measured. Second, balancing the pace at which different tasks are learned. The relative inverse training rate $r_i\left(t\right)$ is used to this end. When the relative inverse training rate $r_i\left(t\right)$ increases, the gradient magnitude $G_i^W\left(t\right)$ for task $i$ should increase as well to stimulate the task to train more quickly. GradNorm tackles both objectives by minimizing the following loss
\begin{equation}
\label{eq: gradnorm}    
\left\vert G_i^W \left(t\right) - \bar{G}^W \left(t\right) \cdot r_i\left(t\right) \right\vert.
\end{equation}
Remember that, the gradient magnitude $G_i^W(t)$ for task $i$ depends on the weighted single-task loss $w_i(t)\cdot L_i(t)$. As a result, the objective in Equation~\ref{eq: gradnorm} can be minimized by adjusting the task-specific weights $w_i$. In practice, during training these task-specific weights are updated in every iteration using backpropagation. After every update, the task-specific weights $w_i\left(t\right)$ are re-normalized in order to decouple the learning rate from the task-specific weights.

Note that, calculating the gradient magnitude $G_i^W\left(t\right)$ requires a backward pass through the task-specific layers of every task $i$. However, savings on computation time can be achieved by considering the task gradient magnitudes only w.r.t. the weights in the last shared layer. 

Different from uncertainty weighting, GradNorm does not take into account the task-dependent uncertainty to re-weight the task-specific losses. Rather, GradNorm tries to balance the pace at which tasks are learned, while avoiding gradients of different magnitude. 

\subsubsection{Dynamic Weight Averaging}
\label{subsubsec: dynamic_weight_averaging}

Similarly to GradNorm, Liu et al.~\cite{liu2019end} proposed a technique, termed \textbf{Dynamic Weight Averaging} (DWA), to balance the pace at which tasks are learned. Differently, DWA only requires access to the task-specific loss values. This avoids having to perform separate backward passes during training in order to obtain the task-specific gradients. In DWA, the task-specific weight $w_i$ for task $i$ at step $t$ is set as 
\begin{equation}
\label{eq: dynamic_weight_averaging}
w_i\left(t\right) = \frac{N \exp \left(r_i \left( t-1 \right)/T \right)}{\sum_n \exp \left(r_n \left(t-1\right)/T\right)}, r_n\left(t-1\right) = \frac{L_n\left(t-1\right)}{L_n\left(t-2\right)},
\end{equation}
with $N$ being the number of tasks. The scalars $r_n\left(\cdot\right)$ estimate the relative descending rate of the task-specific loss values $L_n$. The temperature $T$ controls the softness of the task weighting in the softmax operator. When the loss of a task decreases at a slower rate compared to other tasks, the task-specific weight in the loss is increased.

Note that, the task-specific weights $w_i$ are solely based on the rate at which the task-specific losses change. Such a strategy requires to balance the overall loss magnitudes beforehand, else some tasks could still overwhelm the others during training. GradNorm avoids this problem by balancing both the training rates and the gradient magnitudes through a single objective (see Equation~\ref{eq: gradnorm}).

\subsubsection{Dynamic Task Prioritization}
\label{subsubsec: dynamic_task_prioritization}

The task balancing techniques in Sections~\ref{subsubsec: uncertainty_weighting}-\ref{subsubsec: dynamic_weight_averaging} opted to optimize the task-specific weights $w_i$ as part of a Gaussian likelihood objective~\cite{kendall2018multi}, or in order to balance the pace at which the different tasks are learned~\cite{liu2019end,chen2018gradnorm}. In contrast, \textbf{Dynamic Task Prioritization} (DTP)~\cite{guo2018dynamic} opted to prioritize the learning of 'difficult' tasks by assigning them a higher task-specific weight. The motivation is that the network should spend more effort to learn the 'difficult' tasks. Note that, this is opposed to uncertainty weighting, where a higher weight is assigned to the 'easy' tasks. We hypothesize that the two techniques do not necessarily conflict, but uncertainty weighting seems better suited when tasks have noisy labeled data, while DTP makes more sense when we have access to clean ground-truth annotations.  

To measure the task difficulty, one could consider the progress on every task using the loss ratio $\tilde{L}_i(t)$ defined by GradNorm. However, since the loss ratio depends on the initial loss $L_i(0)$, its value can be rather noisy and initialization dependent. Furthermore, measuring the task progress using the loss ratio might not accurately reflect the progress on a task in terms of qualitative results. Therefore, DTP proposes the use of key performance indicators (KPIs) to quantify the difficulty of every task. In particular, a KPI $\kappa_i$ is selected for every task $i$, with $0 < \kappa_i < 1$. The KPIs are picked to have an intuitive meaning, e.g. accuracy for classification tasks. For regression tasks, the prediction error can be thresholded to obtain a KPI that lies between 0 and 1. Further, we define a task-level focusing parameter $\gamma_i \geq 0 $ that allows to adjust the weight at which easy or hard tasks are down-weighted. DTP sets the task-specific weight $w_i$ for task $i$ at step $t$ as
\begin{equation}
\label{eq: dynamic_task_priorization}
w_i\left(t\right) =  -\left( 1 - \kappa_i \left(t\right)\right)^{\gamma_i} \log \kappa_i \left(t\right).
\end{equation}
Note that, Equation~\ref{eq: dynamic_task_priorization} employs a focal loss expression~\cite{lin2017focal} to down-weight the task-specific weights for the 'easy' tasks. In particular, as the value for the KPI $\kappa_i$ increases, the weight $w_i$ for task $i$ is being reduced.

DTP requires to carefully select the KPIs. For example, consider choosing a threshold to measure the performance on a regression task. Depending on the threshold's value, the task-specific weight will be higher or lower during training. We conclude that the choice of the KPIs in DTP is not determined in a straightforward manner. Furthermore, similar to DWA, DTP requires to balance the overall magnitude of the loss values beforehand. After all, Equation~\ref{eq: dynamic_task_priorization} does not take into account the loss magnitudes to calculate the task-specific weights. As a result, DTP still involves manual tuning to set the task-specific weights. 

\subsubsection{MTL as Multi-Objective Optimization}
\label{subsubsec: mgda_ub}

A global optimum for the multi-task optimization objective in Equation~\ref{eq: mtl_loss} is hard to find. Due to the complex nature of this problem, a certain choice that improves the performance for one task could lead to performance degradation for another task. The task balancing methods discussed beforehand try to tackle this problem by setting the task-specific weights in the loss according to some heuristic. Differently, Sener and Koltun~\cite{sener2018multi} view MTL as a multi-objective optimization problem, with the overall goal of finding a Pareto optimal solution among all tasks. 

In MTL, a Pareto optimal solution is found when the following condition is satisfied: the loss for any task can be decreased without increasing the loss on any of the other tasks. A \textbf{multiple gradient descent algorithm} (MGDA)~\cite{desideri2012multiple} was proposed in~\cite{sener2018multi} to find a Pareto stationary point. In particular, the shared network weights are updated by finding a common direction among the task-specific gradients. As long as there is a common direction along which the task-specific losses can be decreased, we have not reached a Pareto optimal point yet. An advantage of this approach is that since the shared network weights are only updated along common directions of the task-specific gradients, conflicting gradients are avoided in the weight update step.

Lin et al.~\cite{lin2019pareto} observed that MGDA only finds one out of many Pareto optimal solutions. Moreover, it is not guaranteed that the obtained solution will satisfy the users' needs. To address this problem, they generalized MGDA to generate a set of well-representative Pareto solutions from which a preferred solution can be selected. So far, however, the method was only applied to small-scale datasets (e.g. Multi-MNIST).

\subsubsection{Discussion}
\label{subsubsec: task_balancing_discussion}

\begin{table*}[t]
\caption{A qualitative comparison between task balancing techniques. First, we consider whether a method balances the loss magnitudes (\textbf{Balance Magnitudes}) and/or the pace at which tasks are learned (\textbf{Balance Learning}). Second, we show what tasks are prioritized during the training stage (\textbf{Prioritize}). Third, we show whether the method requires access to the task-specific gradients (\textbf{Grads Required}). Fourth, we consider whether the tasks gradients are enforced to be non-conflicting (\textbf{Non-Competing Grads}). Finally, we consider whether the proposed method requires additional tuning, e.g. manually choosing the weights, KPIs, etc. (\textbf{No Extra Tuning}). For a detailed description of each technique see Section~\ref{sec: optimization}.}
\label{tab: task_balancing}
\begin{center}
\begin{tabular}{lccccccl}
\toprule
\textbf{Method} & \textbf{\multirowcell{2}[0pt][l]{Balance \\ Magnitudes}} & \textbf{\multirowcell{2}[0pt][l]{Balance \\ Learning}} &
\textbf{Prioritize} & 
\textbf{\multirowcell{2}[0pt][l]{Grads \\ Required}} &
\textbf{\multirowcell{2}[0pt][l]{Non-Competing \\ Grads}} &
\textbf{\multirowcell{2}[0pt][l]{No Extra \\ Tuning }} & 
\textbf{Motivation} \\
& & & & & & \\
\midrule
Uncertainty~\cite{kendall2018multi} & $\checkmark$ & & Low Noise & & &  $\checkmark$ & Homoscedastic uncertainty \\
GradNorm~\cite{chen2018gradnorm} & $\checkmark$ & $\checkmark$ & & $\checkmark$ & & $\checkmark$ &  Balance learning and magnitudes \\
DWA~\cite{liu2019end} & & $\checkmark$ & & & & &  Balance learning \\
DTP~\cite{guo2018dynamic} & & & Difficult & & & & Prioritize difficult tasks \\
MGDA~\cite{sener2018multi} & $\checkmark$ & $\checkmark$ & & $\checkmark$ & $\checkmark$ & $\checkmark$ & Pareto Optimum \\
\bottomrule
\end{tabular}
\end{center}
\end{table*}

In Section~\ref{subsec: task_balancing}, we described several methods for balancing the influence of each task when training a multi-task network. Table~\ref{tab: task_balancing} provides a qualitative comparison of the described methods. We summarize some conclusions below. (1) We find discrepancies between these methods, e.g. uncertainty weighting assigns a higher weight to the 'easy' tasks, while DTP advocates the opposite. The latter can be attributed to the experimental evaluation of the different task balancing strategies that was often done in the literature using different datasets or task dictionaries. We suspect that an appropriate task balancing strategy should be decided for each case individually. (2) We also find commonalities between the described methods, e.g. uncertainty weighting, GradNorm and MGDA opted to balance the loss magnitudes as part of their learning strategy. In Section~\ref{subsec: experiments_task_balancing}, we provide extensive ablation studies under more common datasets or task dictionaries to verify what task balancing strategies are most useful to improve the multi-tasking performance, and under which circumstances. (3) A number of works (e.g. DWA, DTP) still require careful manual tuning of the initial hyperparameters, which can limit their applicability when dealing with a larger number of tasks. 

\subsection{Other Approaches}
\label{subsec: task_balancing_other}

The task balancing works in Section~\ref{subsec: task_balancing} can be plugged into most existing multi-task architectures to regulate the task learning. Another group of works also tried to regulate the training of multi-task networks, albeit for more specific setups. We touch upon several of these approaches here. Note that, some of these concepts can be combined with task balancing strategies too. 

Zhao et al.~\cite{zhao2018modulation} empirically found that tasks with gradients pointing in the opposite direction can cause the destructive interference of the gradient. This observation is related to the update rule in Equation~\ref{eq: update_rule}. They proposed to add a modulation unit to the network in order to alleviate the competing gradients issue during training. 

Liu et al.~\cite{liu2017adversarial} considered a specific multi-task architecture where the feature space is split into a shared and a task-specific part. They argue that the shared features should contain more common information, and no information that is specific to a particular task only. The network weights are regularized by enforcing this prior. More specifically, an adversarial approach is used to avoid task-specific features from creeping into the shared representation. Similarly,~\cite{sinha2018gradient,maninis2019attentive} added an adversarial loss to the single-task gradients in order to make them statistically indistinguishable from each other in the shared parts of the network.

Chen et al.~\cite{chen2020just} proposed gradient sign dropout, a modular layer that can be plugged into any network with multiple gradient signals. Following MGDA, the authors argue that conflicts in the weight update arise when the gradient values of the different learning signals have opposite signs. Gradient sign dropout operates by choosing the sign of the gradient based on the distribution of the gradient values, and masking out the gradient values with opposite sign. It is shown that the method has several desirable properties, and increases performance and robustness compared to competing works.

Finally, some works relied on heuristics to balance the tasks. Sanh et al.~\cite{sanh2019hierarchical} trained the network by randomly sampling a single task for the weight update during every iteration. The sampling probabilities were set proportionally to the available amount of training data for every task. Raffel et al.~\cite{raffel2019exploring} used temperature scaling to balance the tasks. So far, however, both procedures were used in the context of natural language processing. 

\section{Experiments}
\label{sec: experiments}
This section provides an extensive comparison of the previously discussed methods. First, we describe the experimental setup in Section~\ref{subsec: experiments_setup}. We cover the used datasets, methods, evaluation criteria and training setup, so the reader can easily give interpretation to the obtained results. Section~\ref{subsec: experiments_overview} presents a general overview of the results, allowing us to identify several overall trends. Section~\ref{subsec: experiment_architectures} compares the MTL architectures in more detail, while the task balancing strategies are considered in Section~\ref{subsec: experiments_task_balancing}. We refer the reader to the supplementary materials for qualitative results. 

\subsection{Experimental Setup}
\label{subsec: experiments_setup}

\subsubsection{Datasets}
\label{subsubsec: experiments_datasets}
Our experiments were conducted on two popular dense-labeling benchmarks, i.e. NYUD-v2~\cite{silberman2012indoor} and PASCAL~\cite{everingham2010pascal}. We selected the datasets to provide us with a diverse pair of settings, allowing us to scrutinize the advantages and disadvantages of the methods under consideration. We additionally took into account what datasets were used in the original works. Both datasets are described in more detail below.

The \textbf{PASCAL} dataset~\cite{everingham2010pascal} is a popular benchmark for dense prediction tasks. We use the split from PASCAL-Context~\cite{chen2014detect} which has annotations for semantic segmentation, human part segmentation and semantic edge detection. Additionally, we consider the tasks of surface normals prediction and saliency detection. The annotations were distilled by~\cite{maninis2019attentive} using pre-trained state-of-the-art models~\cite{bansal2017pixelnet,chen2018encoder}. The optimal dataset F-measure (\emph{odsF})~\cite{martin2004learning} is used to evaluate the edge detection task. The semantic segmentation, saliency estimation and human part segmentation tasks are evaluated using mean intersection over union (\emph{mIoU}). We use the mean error (\emph{mErr}) in the predicted angles to evaluate the surface normals. 

The \textbf{NYUD-v2} dataset~\cite{silberman2012indoor} considers indoor scene understanding. The dataset contains 795 train and 654 test images annotated for semantic segmentation and monocular depth estimation. Other works have also considered surface normal prediction~\cite{xu2018pad,zhang2019pattern,maninis2019attentive} and semantic edge detection~\cite{xu2018pad,maninis2019attentive} on the NYUD-v2 dataset. The annotations for these tasks can be directly derived from the semantic and depth ground truth. In this work we focus on the semantic segmentation and depth estimation tasks. We use the mean intersection over union (\emph{mIoU}) and root mean square error (\emph{rmse}) to evaluate the semantic segmentation and depth estimation task respectively. 

Table~\ref{tab: datasets} gives an overview of the used datasets. We marked tasks for which the annotations were obtained through distillation with an asterik. 

\begin{table*}[t]
\caption{MTL benchmarks used in the experiments section. Distilled task labels are marked with *.}
\label{tab: datasets}
\begin{center}
\begin{tabular}{lcccccccc}
\toprule
\textbf{Dataset} & \textbf{\# Train} & \textbf{\# Test} & \textbf{Segmentation} & \textbf{Depth} & \textbf{Human Parts} & \textbf{Normals} & \textbf{Saliency} & \textbf{Edges} \\
\midrule
PASCAL~\cite{everingham2010pascal} & 10,581 & 1,449 & $\checkmark$ &  & $\checkmark$ & $\checkmark$* & $\checkmark$* & $\checkmark$ \\
NYUD-v2~\cite{silberman2012indoor} & 795 & 654 & $\checkmark$ & $\checkmark$ & & & & \\ 
\bottomrule
\end{tabular}
\end{center}
\end{table*}

\subsubsection{Evaluation Criterion}
\label{subsubsec: experiments_evaluation}
In addition to reporting the performance on every individual task, we include a single-number performance metric for the multi-task models. Following prior work~\cite{maninis2019attentive,vandenhende2019branched,vandenhende2020mti,bruggemann2020automated}, we define the multi-task performance $\Delta_{MTL}$ of a multi-task learning model $m$ as the average per-task drop in performance w.r.t. the single-task baseline $b$:
\begin{equation}
\label{eq: evaluation_metric}
\Delta_{MTL} = \frac{1}{T} \sum_{i=1}^{T} \left(-1\right)^{l_i} \left(M_{m,i} - M_{b,i} \right) / M_{b,i},
\end{equation}
where $l_i = 1$ if a lower value means better performance for metric $M_i$ of task $i$, and 0 otherwise. The single-task performance is measured for a fully-converged model that uses the same backbone network only to perform that task. To achieve a fair comparison, all results were obtained after performing a grid search on the hyperparameters. This ensures that every model is trained with comparable amounts of finetuning. We refer to Section~\ref{subsubsec: experiments_train_setup} for more details on our training setup. 

The MTL performance metric does not account for the variance when different hyperparameters are used. To address this, we analyze the influence of the used hyperparameters on NYUD-v2 with performance profiles. Finally, in addition to a performance evaluation, we also include the model resources, i.e. number of parameters and FLOPS, when comparing the multi-task architectures. 

\subsubsection{Compared Methods}
\label{subsubsec: experiments_methods}
Table~\ref{tab: experiments_methods} summarizes the models and task balancing strategies used in our experiments. We consider the following encoder-focused architectures from Section~\ref{subsec: encoder_focused_architectures} on NYUD-v2 and PASCAL: MTL baseline with shared encoder and task-specific decoders, cross-stitch networks~\cite{misra2016cross}, NDDR-CNN~\cite{gao2019nddr} and MTAN~\cite{liu2019end}. We do not include branched MTL networks, since this collection of works is mainly situated in the domain of Neural Architecture Search, and focuses on finding a MTL solution that fits a specific computational budget constraint. We refer to the corresponding papers~\cite{lu2017fully,vandenhende2019branched,bruggemann2020automated,guo2020learning} for a concrete experimental analysis on this subject. All compared models use a ResNet~\cite{he2016deep} encoder with dilated convolutions~\cite{chen2018encoder}. We use the ResNet-50 variant for our experiments on NYUD-v2. On PASCAL, we use the shallower ResNet-18 model due to GPU memory constraints. The task-specific heads use an Atrous Spatial Pyramid Pooling (ASPP)~\cite{chen2018encoder} module.

Furthermore, we cover the following decoder-focused approaches from Section~\ref{subsec: decoder_focused_architectures}: JTRL~\cite{zhang2018joint}, PAP-Net~\cite{zhang2019pattern}, PAD-Net~\cite{xu2018pad} and MTI-Net~\cite{vandenhende2020mti}. Note that a direct comparison between all models is not straightforward. There are several reasons behind this. First, MTI-Net operates on top of a multi-scale feature representation of the input image, which assumes a multi-scale backbone, unlike the other approaches that were originally designed with a single-scale backbone network in mind. Second, JTRL was strictly designed for a pair of tasks, without any obvious extension to the MTL setting. Finally, PAP-Net behaves similarly to PAD-Net, but operates on the pixel affinities to perform the multi-modal distillation through a recursive diffusion process. 

Based on these observations, we organize the experiments as follows. On NYUD-v2, we consider PAD-Net, PAP-Net and JTRL in combination with a ResNet-50 backbone. This facilitates an apples-to-apples comparison between the encoder- and decoder-focused approaches that operate on top of a single-scale feature extractor. We draw a separate comparison between MTI-Net and PAD-Net on NYUD-v2 using a multi-scale HRNet-18 backbone~\cite{sun2019deep}. Finally, we repeat the comparison between PAD-Net and MTI-Net on PASCAL using the multi-scale HRNet-18 backbone to verify how well the decoder-focused approaches handle the larger and more diverse task dictionary. 
In addition to the MTL architectures, we also compare the task balancing techniques from Section~\ref{subsec: task_balancing}. We analyze the use of fixed weights, uncertainty weighting~\cite{kendall2018multi}, GradNorm~\cite{chen2018gradnorm}, DWA~\cite{liu2019end} and MGDA~\cite{sener2018multi}. We did not include DTP~\cite{guo2018dynamic}, as this technique requires to define additional key performance indicators (see Section~\ref{subsubsec: dynamic_task_prioritization}). The task balancing methods are evaluated in combination with the MTL baseline models that use a single-scale ResNet backbone. We do not consider how the task balancing techniques interact with other MTL architectures. This choice stems from the following observations. First, GradNorm and MGDA were specifically designed with the vanilla hard parameter sharing model in mind (i.e. MTL baseline). Second, uncertainty weighting and DWA re-weigh the tasks based on the task-specific losses. Since the loss values depend more on the used loss functions, and less on the used architectures, we expect these methods to result in similar task weights when plugging in a different model. 

\begin{table*}[ht]
    \centering
    \caption{Overview of our experiments on NYUD-v2 and PASCAL. We indicate the used backbones and optimization strategies for every model.}
    \label{tab: experiments_methods}
    \begin{tabular}{lcclllc}
    \toprule
    \textbf{Model}   &   \multicolumn{2}{c}{\textbf{Datasets}}    &  \multicolumn{2}{c}{\textbf{Backbone}}     &   \textbf{Optimization}\\
            &   \textbf{NYUD-v2} & \textbf{PASCAL} & \textbf{NYUD-v2} & \textbf{PASCAL} &     \\
    \midrule
    MTL Baseline & $\checkmark$ & $\checkmark$ & ResNet-50/HRNet-18 & ResNet-18/HRNet-18 & Uniform, Fixed (Grid Search), Uncertainty, \\
     & & & & & GradNorm, DWA, MGDA \\
    Cross-Stitch &   $\checkmark$  & $\checkmark$    & ResNet-50 & ResNet-18 & Fixed (Grid Search) \\
    NDDR-CNN &   $\checkmark$  & $\checkmark$    & ResNet-50 & ResNet-18 & Fixed (Grid Search) & \\
    MTAN &   $\checkmark$  & $\checkmark$    & ResNet-50 & ResNet-18 & Fixed (Grid Search) & \\
    \midrule
    JTRL & $\checkmark$  & & ResNet-50 & - & Fixed (Grid Search) \\
    PAP-Net & $\checkmark$  & & ResNet-50 & - & Fixed (Grid Search) \\
    PAD-Net & $\checkmark$  & $\checkmark$    & ResNet-50/HRNet-18 & HRNet-18 & Fixed (Grid Search) \\
    MTI-Net & $\checkmark$  & $\checkmark$    & HRNet-18 & HRNet-18 & Fixed (Grid Search) \\
    \bottomrule    
    \end{tabular}
\end{table*}

\subsubsection{Training Setup}
\label{subsubsec: experiments_train_setup}
We reuse the loss functions and augmentation strategy from~\cite{vandenhende2020mti}. The MTL models are trained with fixed loss weights obtained from~\cite{maninis2019attentive}, that optimized them through a grid search. All experiments are performed using pre-trained ImageNet weights. The optimizer, learning rate and batch size are optimized with a grid search procedure to ensure a fair comparison across all compared approaches. More specifically, we test batches of size 6 and 12, and Adam (LR=$\left\{\text{1e-4, 5e-4}\right\}$) vs stochastic gradient descent with momentum 0.9 (LR=$\left\{\text{1e-3,5e-3,1e-2,5e-2}\right\}$). This accounts for 12 hyperparameter settings in total (see overview in Table~\ref{tab: grid_search}). A poly learning rate scheduler is used. The number of total epochs is set to 60 for PASCAL and 100 for NYUD-v2. We include weight decay regularization of 1e-4. Any remaining hyperparameters are set in accordance with the original works. 

\begin{table}
\centering
\caption{Overview of the used hyperparameter settings in our grid search procedure.}
\label{tab: grid_search}
\begin{tabular}{lll}
\toprule
\textbf{Optimizer} & \textbf{Batch size} & \textbf{Learning rate} \\
\midrule
SGD ($momentum=0.9$) & $\left\{\text{6, 12}\right\}$ & $\left\{\text{1e-3, 5e-3, 1e-2, 5e-2}\right\}$ \\
Adam & $\left\{\text{6, 12}\right\}$ & $\left\{\text{1e-4, 5e-4}\right\}$\\
\bottomrule
\end{tabular}
\end{table}

\begin{table*}[ht]
\centering
\caption{Comparison of deep architectures and optimization strategies for MTL on NYUD-v2 and PASCAL. Models that use the same backbone can be put in direct comparison with each other. $\Delta_{MTL}$ indicates the average relative improvement over the single-task baselines (see Equation~\ref{eq: evaluation_metric}). For more details about the experimental setup visit Section~\ref{subsec: experiments_setup}.}
\label{tab: experiments_all}
\begin{subtable}[t]{\linewidth}
\centering
\caption{Comparison of deep MTL architectures for MTL on the NYUD-v2 validation set.}
\label{tab: nyud_models}
    \begin{tabular}{cllccccc}
    \toprule
    & \textbf{Backbone} & \textbf{Model} & \textbf{FLOPS (G)} & \textbf{Params (M)} & \textbf{Seg. (IoU)} $\mathbf{\uparrow}$ & \textbf{Depth (rmse)} $\mathbf{\downarrow}$ & $\mathbf{\Delta_{MTL}(\%)} \uparrow$\\
    \midrule
    \multirow{2}{*}{\shortstack{\textbf{Single}\\\textbf{Task}}} & ResNet-50  & & 192 & 80 & 43.9 & 0.585 & + 0.00\\
    & \textit{HRNet-18} & & 11 & 8 & 35.3 & 0.648 & + 0.00 \\
    \midrule 
    \multirow{5}{*}{\shortstack{\textbf{Encoder}\\\textbf{Focused}\\\textbf{MTL}}} & \multirow{4}{*}{ResNet-50} & MTL Baseline & 133 & 56 & 44.4 & 0.587 & + 0.41\\
    & & MTAN  & 197 & 72 & 45.0 & 0.584 & + 1.32\\
    & & Cross-Stitch & 192 & 80 & 44.2 & 0.570 & + 1.61\\
    & & NDDR-CNN & 207 & 102 & 44.2 & 0.573 & + 1.38 \\
    & \textit{HRNet-18} & \textit{MTL Baseline} & 6 & 4 & 33.9 & 0.636 & - 1.09\\
    \midrule    
    \multirow{5}{*}{\shortstack{\textbf{Decoder}\\\textbf{Focused}\\\textbf{MTL}}} & \multirow{3}{*}{ResNet-50} & JTRL & 660 & 295 & 46.4 & 0.501 & + 10.02 \\
    & & PAP-Net & 4800 & 52 & 50.4 & 0.530 & + 12.10\\
    & & PAD-Net (Single-Scale) & 256 & 52 & 50.2 & 0.582 & + 7.43\\
    & \multirow{2}{*}{\textit{HRNet-18}} & \textit{PAD-Net (Multi-Scale)} & 82 & 12 & 36.0 & 0.630 & + 2.38 \\
    & & \textit{MTI-Net} & 16 & 27 & 38.6 & 0.593 & + 8.95 \\
    \bottomrule
    \end{tabular}
\end{subtable}

\vspace{0.25cm}

\begin{subtable}[t]{\linewidth}
\centering
\caption{Comparison of deep MTL architectures for MTL on the PASCAL validation set.}
\label{tab: pascal_models}
    \begin{tabular}{cllcccccccc}
    \toprule
    & \textbf{Backbone} & \textbf{Model} & \textbf{FLOPS} & \textbf{Params} & \textbf{Seg.}& \textbf{H. Parts} & \textbf{Norm.} & \textbf{Sal.} & \textbf{Edge} & $\mathbf{\Delta_{MTL}}$\\
    & & & \textbf{(G)} & \textbf{(M)} &  \textbf{(IoU)} $\mathbf{\uparrow}$ & \textbf{(IoU)} $\mathbf{\uparrow}$ & \textbf{(mErr)} $\mathbf{\downarrow}$ & \textbf{(IoU)} $\mathbf{\uparrow}$ & \textbf{(odsF)} $\mathbf{\uparrow}$ & $\mathbf{(\%) \uparrow}$ \\
    \midrule
    \multirow{2}{*}{\shortstack{\textbf{Single}\\\textbf{Task}}} & ResNet-18 & & 167 & 80 & 66.2 & 59.9 & 13.9 & 66.3 & 68.8 & + 0.00 \\
    & \textit{HRNet-18} & & 24 & 20 & 59.5 & 61.4 & 14.0 & 67.3 & 72.6 & + 0.00 \\
    \midrule
    \multirow{5}{*}{\shortstack{\textbf{Encoder}\\\textbf{Focused}\\\textbf{MTL}}} & \multirow{4}{*}{\shortstack{ResNet-18}} & MTL Baseline & 71 & 35 & 63.8 & 58.6 & 14.9 & 65.1 & 69.2 & - 2.86\\
    & & MTAN & 80 & 37 & 63.7 & 58.9 & 14.8 & 65.4 & 69.6 & - 2.39 \\
    & & Cross-Stitch & 167 & 80 & 66.1 & 60.6 & 13.9 & 66.8 & 69.9 & + 0.60 \\
    & & NDDR-CNN & 187 & 88 & 65.4 & 60.5 & 13.9 & 66.8 & 69.8 & + 0.39\\
    & \textit{HRNet-18} & \textit{MTL Baseline} & 7 & 4 & 54.3 & 59.3 & 14.8 & 65.5 & 71.7 & - 4.38\\
    \midrule    
    \multirow{3}{*}{\shortstack{\textbf{Decoder}\\\textbf{Focused}\\\textbf{MTL}}} & ResNet-18 & PAD-Net & 36 & 32 & 63.2 & 59.3 & 15.2 & 64.3 & 60.2 & - 5.62 \\
     & \multirow{2}{*}{\shortstack{\textit{HRNet-18}}} & \textit{PAD-Net} & 212 & 29 & 53.6 & 59.6 & 15.3 & 65.8 & 72.5 & - 4.41 \\
    & & \textit{MTI-Net} & 15 & 24 & 64.3 & 62.1 & 14.8 & 68.0 & 73.4 & + 1.13 \\
    \bottomrule
    \end{tabular}
\end{subtable}

\vspace{0.25cm}

\begin{subtable}[t]{0.38\linewidth}
\centering
\caption{Comparison of optimization strategies for MTL on the NYUD-v2 validation set. The MTL baseline model with ResNet-50 backbone is used.}
\label{tab: nyud_optimization}
    \begin{tabular}{lccc}
    \toprule
    \textbf{Method} & \textbf{Seg. } & \textbf{Depth} &  $\mathbf{\Delta_{MTL}}$ \\
    & \textbf{(IoU)} $\mathbf{\uparrow}$ & \textbf{(rmse)} $\mathbf{\downarrow}$ &(\%)$\mathbf{\uparrow}$\\
    \midrule
    Single-Task & 43.9 & 0.585 & + 0.00 \\
    \midrule
    Fixed (Grid S.) & \textbf{44.4} & 0.587 & + 0.41 \\
    Uncertainty & 44.0 & 0.590 & - 0.23 \\
    DWA & 44.1 & 0.591 & - 0.28 \\
    GradNorm & 44.2 & \textbf{0.581} & \textbf{+ 1.45} \\
    MGDA & 43.2 & 0.576 & + 0.02 \\
    \bottomrule
    \end{tabular}
\end{subtable}
\hspace{\fill}
\begin{subtable}[t]{0.6\linewidth}
\centering
\caption{Comparison of optimization strategies for MTL on the PASCAL validation set. The MTL baseline model with ResNet-18 backbone is used.}
\label{tab: pascal_optimization}
    \begin{tabular}{lccccccc}
    \toprule
    \textbf{Method} & \textbf{Seg.}& \textbf{H. Parts} & \textbf{Norm.} & \textbf{Sal.} & \textbf{Edge} & $\mathbf{\Delta_{MTL}}$\\
    &  \textbf{(IoU)} $\mathbf{\uparrow}$ & \textbf{(IoU)} $\mathbf{\uparrow}$ & \textbf{(mErr)} $\mathbf{\downarrow}$ & \textbf{(IoU)} $\mathbf{\uparrow}$ & \textbf{(odsF)} $\mathbf{\uparrow}$ & $\mathbf{(\%) \uparrow}$ \\
    \midrule
    Single-Task & \textbf{66.2} & \textbf{59.9} & \textbf{13.9} & \textbf{66.3} & 68.8 & \textbf{+ 0.00} \\
    \midrule
    Uniform & 65.5 & 59.5 & 15.8 & 64.1 & 67.9 & - 3.99 \\
    Fixed (Grid S.) & 63.8 & 58.6 & 14.9 & 65.1 & \textbf{69.2} & - 2.86 \\
    Uncertainty & 65.4 & 59.2 & 16.5 & 65.6 & 68.6 & - 4.60 \\
    DWA & 63.4 & 58.9 & 14.9 & 65.1 & 69.1 & - 2.94 \\
    GradNorm & 64.7 & 59.0 & 15.4 & 64.5 & 67.0 & - 3.97 \\
    MGDA & 64.9 & 57.9 & 15.6 & 62.5 & 61.4 & -6.81 \\
    \bottomrule
    \end{tabular}
\end{subtable}

\end{table*}

\subsection{Overview}
\label{subsec: experiments_overview}
Table~\ref{tab: experiments_all} provides an overview of the results on NYUD-v2 and PASCAL. The MTL architectures are shown in Tables~\ref{tab: nyud_models} and~\ref{tab: pascal_models}. A direct comparison can be made between architectures that rely on the same backbone. The task balancing strategies are analyzed in Tables~\ref{tab: nyud_optimization} and~\ref{tab: pascal_optimization}. We identify several trends from the results. 

\noindent\textbf{Single-Task vs Multi-Task.}
We compare the encoder- and decoder-focused MTL models against their single-task counterparts on NYUD-v2 and PASCAL in Tables~\ref{tab: nyud_models}-~\ref{tab: pascal_models}. MTL can offer several advantages relative to single-task learning, that is, smaller memory footprint, reduced number of calculations, and improved performance. However, few models are able to deliver on this potential to its full extent. For example, JTRL improves the performance on the segmentation and depth estimation tasks on NYUD-v2, but requires more resources. Differently, the processing happens more efficiently when using the MTL baseline on PASCAL, but the performance declines too. MTI-Net constitutes an exemption from this rule. In particular, the performance increases on all tasks, except for normals, while the computational overhead is limited. Note, in this specific case, the relative increase in parameters and FLOPS can be attributed to the use of a shallow backbone network. 

\noindent\textbf{Influence of Task Dictionary.}
We study the influence of the task dictionary (i.e. size and diversity) by comparing the results on NYUD-v2 against PASCAL (see Table~\ref{tab: nyud_models} vs Table~\ref{tab: pascal_models}). On NYUD-v2, we consider the tasks of semantic segmentation and depth estimation. This pair of tasks is strongly related~\cite{xu2018pad,zhang2019pattern,zhou2020pattern}, since both semantic segmentation and depth estimation reveal similar characteristics about a scene, such as the layout and object shapes or boundaries. Differently, PASCAL includes a larger and more diverse task dictionary. 

On NYUD-v2, MTL proves an effective strategy for jointly tackling segmentation and depth estimation. In particular, most MTL models outperform the set of single-task networks (see Table~\ref{tab: nyud_models}). Similar results have been reported for other pairs of well-correlated tasks, e.g. depth and flow estimation~\cite{zou2018df}, detection and classification~\cite{girshick2015fast,ren2015faster}, detection and segmentation~\cite{dvornik2017blitznet,he2017mask}.

Differently, most existing models fail to outperform their single-task equivalents on PASCAL (see Table~\ref{tab: pascal_models}). For example, the improvements reported by the encoder-focused approaches are usually limited to a few isolated tasks, while a decline in performance is observed for the other tasks. Jointly tackling a large and diverse task dictionary proves challenging, as also noted by~\cite{vandenhende2019branched,maninis2019attentive}. 

\noindent\textbf{Architecture vs Optimization.}
The effect of designing a better MTL architecture is compared against the use of a better task balancing strategy (see Tables~\ref{tab: nyud_models}-\ref{tab: pascal_models} vs~\ref{tab: nyud_optimization}-\ref{tab: pascal_optimization}). We find that the use of a better MTL architecture is usually more helpful to improve the performance in MTL. Similar observations were made by prior works~\cite{liu2019end, maninis2019attentive}. 

\noindent\textbf{Encoder- vs Decoder-Focused Models.}
We compare the encoder-focused models against the decoder-focused models on NYUD-v2 and PASCAL in Tables~\ref{tab: nyud_models}-~\ref{tab: pascal_models}. First, we find that the decoder-focused architectures generally outperform the encoder-focused ones in terms of multi-task performance. We argue that each architecture paradigm serves different purposes. The encoder-focused architectures aim to learn richer feature representations of the image by sharing information during the encoding process. Differently, the decoder-focused ones focus on improving dense prediction tasks by repeatedly refining the predictions through cross-task interactions. Since the interactions take place near the output of the network, they allow for a better alignment of common cross-task patterns, which in turn, greatly boost the performance. Based on their complementary behavior, we hope to see both paradigms integrated in future works. 

Second, we focus on the encoder- and decoder-focused models that use an identical ResNet-50 backbone on NYUD-v2 in Table~\ref{tab: nyud_models}. The decoder-focused models report higher performance, but consume a large number of FLOPS. The latter is due to repeatedly predicting the task outputs at high resolution scales. On the other hand, except for JTRL, the decoder-focused models have a smaller memory footprint compared to the encoder-focused ones. We argue that the decoder-focused approaches parameterize the task interactions more efficiently. This can be understood as follows. The task features before the final layer are disentangled according to the structure of the output task. This allows to distill the relevant cross-task patterns with a small number of filter operations. This situation is different for the encoder-focused approaches, where tasks share information in the intermediate layers of the encoder. 

\subsection{Architectures}
\label{subsec: experiment_architectures}
We study the MTL architectures in more detail. Section~\ref{subsubsec: experiments_encoder} compares the encoder-focused architectures, while the decoder-focused ones are discussed in Section~\ref{subsubsec: experiments_decoder}.

\begin{figure*}[ht]
\caption{Performance profile of MTL methods for the semantic segmentation and depth estimation tasks on NYUD-v2. We show the results obtained with different hyperparameter settings in the same color. Bottom-right is better.}
\label{fig: hyperparam}
\begin{subfigure}[t]{0.30\linewidth}
    \centering
    \includegraphics[width=\textwidth]{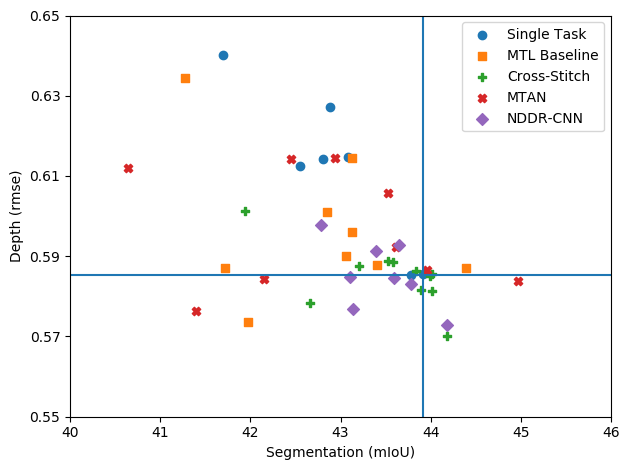}
    \caption{Performance profile of the encoder-focused models.}
    \label{fig: hyperparam_encoder}
\end{subfigure}
\hspace{0.04\linewidth}
\begin{subfigure}[t]{0.30\linewidth}
    \centering
    \includegraphics[width=\textwidth]{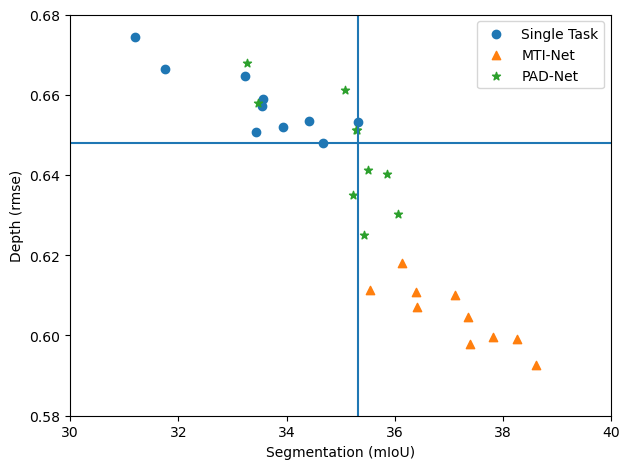}
    \caption{Performance profile of the decoder-focused models.}
    \label{fig: hyperparam_decoder}
\end{subfigure}
\hspace{0.04\linewidth}
\begin{subfigure}[t]{0.30\linewidth}
    \centering
    \includegraphics[width=\textwidth]{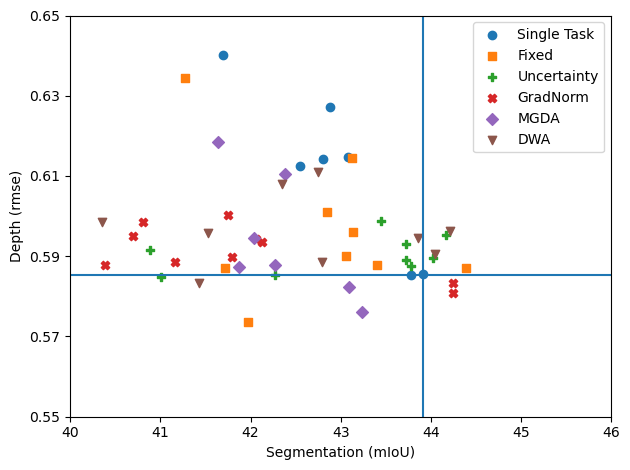}
    \caption{Performance profile of the task balancing techniques.}
    \label{fig: hyperparam_optimization}
\end{subfigure}
\end{figure*}

\subsubsection{Encoder-Focused Architectures}
\label{subsubsec: experiments_encoder}

\noindent\textbf{NYUD-v2.} We analyze the encoder-focused approaches on NYUD-v2 in Table~\ref{tab: nyud_models}. The MTL baseline performs on par with the set of single-task networks, while it reduces the number of parameters and FLOPS. In particular, we observe an increase in performance on the segmentation task ($+0.5$ IoU), and a small decline in performance on the depth estimation task ($+ 0.002$ rmse). Other encoder-focused architectures further improve over the MTL baseline in terms of multi-task performance, but require more parameters and FLOPS. Note that the observed performance gains are of rather small magnitude. We conclude that sharing a ResNet backbone in combination with strong task-specific head units proves a strong baseline for solving a pair of well-correlated dense prediction tasks, like semantic segmentation and depth estimation. 

Furthermore, the cross-stitch network outperforms the NDDR-CNN model both in terms of performance and computational efficiency. Both models follow a similar design, where features from the single-task networks are fused across several encoding layers (see Section~\ref{subsec: encoder_focused_architectures}). The difference lies in the employed feature fusion mechanism: NDDR-CNNs employ non-linear dimensionality reduction to fuse the features, while cross-stitch networks opt for a simple linear combination of the feature channels. Given the more sophisticated feature fusion scheme employed by NDDR-CNNs, one would expect the NDDR-CNN to obtain higher performance. Yet, the opposite is observed in this experiment. We conclude that the feature fusion scheme employed by cross-stitch networks and NDDR-CNNs could benefit from further investigation. 

Finally, no encoder-focused model is significantly outperforming its competitors on the NYUD-v2 benchmark. Therefore, in this case, Multi-Task Attention Networks seem the most favorable choice given their efficient design and good performance. 

\noindent\textbf{PASCAL.} We study again the encoder-focused approaches on the PASCAL dataset in Table~\ref{tab: pascal_models}. In contrast to the results on NYUD-v2, both the MTL baseline and MTAN model report lower performance when compared against the set of single-task models ($-2.86\%$ and $-2.39\%$). There are several possible explanations for this. First, we consider a larger task dictionary. It has been shown~\cite{kokkinos2017ubernet,maninis2019attentive} that a diverse task dictionary is more prone to task interference. Second, the PASCAL dataset contains more labeled examples compared to NYUD-v2. Prior work~\cite{gong2019comparison} observed that the multi-task performance gains can be lower when more annotations are available. Still both models are a useful strategy to reduce the required amount of resources. Depending on the performance requirements, the MTAN model is favored over the MTL baseline or vice versa. 

Furthermore, we observe that cross-stitch networks and NDDR-CNNs can not handle the larger, more diverse task dictionary: the performance improvements are negligible when using these models, while the number of parameters and FLOPS increases. Differently, the MTL baseline and MTAN model are able to strike a better compromise between the multi-task performance and the computation resources that are required. 

\noindent\textbf{Hyperparameters.}
We evaluate the performance of the encoder-focused models when trained with different sets of hyperparameters (see Table~\ref{tab: grid_search}). Figure~\ref{fig: hyperparam_encoder} shows a performance profile for the semantic segmentation and depth estimation tasks on NYUD-v2. Experiments that use the same model, but a different set of hyperparameters, are displayed in identical color. We make several observations regarding the influence of the employed hyperparameters.

First, the performance gains reported by the encoder-focused models are strongly dependent on how well the single-task models were optimized. In particular, the MTL performance gains are significantly enlarged when using a suboptimal set of hyperparameters vs the optimal ones to train the single-task models. We emphasize the importance of carefully training the single-task baselines in MTL. 

Second, the encoder-focused models seem more robust to the used hyperparameters compared to the single-task models. More specifically, when using a less carefully tuned set of hyperparameters, the performance of the single-task models drops faster compared to the encoder-focused MTL models. We will come to a similar conclusion when repeating this experiment for the optimization strategies. We conclude that studying hyperparameter robustness under a single- vs multi-task scenario could be an interesting direction for future work. 

Finally, the performance of cross-stitch networks and NDDR-CNNs is less hyperparameter dependent compared to the MTL baseline and MTAN. The latter pair of models shows much larger performance differences across different hyperparameter settings. 

\noindent\textbf{Discussion.} We compared several encoder-focused architectures on NYUD-v2 and PASCAL. For specific pairs of tasks, e.g. depth estimation and semantic segmentation, we can boost the overall performance through encoder-focused MTL models. However, when considering a large or diverse task dictionary, the performance improvements are limited to a few isolated tasks. In the latter case, MTL still provides a useful strategy to reduce the required amount of computational resources. Notably, there was no encoder-focused model that consistently outperformed the other architectures. Instead, an appropriate MTL model should rather be decided on a per case basis, also taking into account the amount of available computational resources. For example, when performance is crucial, it is advised to use a cross-stitch network, while if the available resources are limited, the MTAN model provides a more viable alternative. 

\subsubsection{Decoder-Focused Architectures}
\label{subsubsec: experiments_decoder}

\noindent\textbf{NYUD-v2.} The results for the decoder-focused models on NYUD-v2 can be found in Table~\ref{tab: nyud_models}. All single-scale decoder-focused architectures report significant gains over the single-task networks on NYUD-v2. PAP-Net achieves the highest multi-task performance ($+12.10\%$), but consumes a large number of FLOPS. This is due to the use of task-affinity matrices, which require to calculate the feature correlation between every pair of pixels in the image. A similar improvement is seen for JTRL ($+10.02\%$). JTRL recursively predicts the two tasks at increasingly higher scales. Therefore, a large number of filter operations is performed on high resolution feature maps, leading to an increase in the computational resources. Opposed to JTRL and PAP-Net, PAD-Net does not come with a significant increase in the number of computations. Yet, we still observe a relatively large improvement over the single-task networks ($+7.43\%$). 

Next, we consider the decoder-focused architectures that used a multi-scale backbone, i.e. HRNet-18. Again, PAD-Net outperforms the single-task networks ($+2.38\%$), but MTI-Net further improves the performance ($+8.95\%$). We conclude that it is beneficial to distill task information at a multitude of scales, instead of a single scale. MTI-Net consumes slightly more parameters and FLOPS compared to the single-task networks. This is due to the use of a rather shallow backbone, i.e. HRNet-18, and a small number of tasks. As a consequence, the overhead introduced by adding the additional layers in MTI-Net is relatively large compared to the resources used by the backbone. PAD-Net sees a large increase in the number of FLOPS compared to MTI-Net. This is due to the fact that PAD-Net performs the multi-modal distillation at a single higher scale (1/4) with $4 \cdot C$ channels, $C$ being the number of backbone channels at a single scale. Instead, MTI-Net performs most of the computations at smaller scales (1/32, 1/16, 1/8), while operating on only $C$ channels at the higher scale (1/4).  

\noindent\textbf{PASCAL.} We analyze the decoder-focused models on PASCAL in Table~\ref{tab: pascal_models}. We see that PAD-Net can not handle the larger task dictionary. The low performance on the semantic edge prediction task can be attributed to not using skip connections in the implemented model. Differently, MTI-Net improves the performance on all tasks, except normals, while requiring fewer flops when compared against the single-task networks. The consistent findings in the larger task dictionary backup our hypothesis about the importance of performing the multi-modal distillation at a multitude of scales (see multi-scale decoder-focused approaches on NYUD-v2). 

\noindent\textbf{Hyperparameters.} Figure~\ref{fig: hyperparam_decoder} shows the performance profile for PAD-Net and MTI-Net on NYUD-v2. We use the same hyperparameters as before (see Table~\ref{tab: grid_search}). The obtained MTL solutions outperform training dedicated single-task models. Furthermore, both PAD-Net and MTI-Net are robust against hyperparameter changes. Even when trained with suboptimal hyperparameters, both models still outperform their single-task counterparts. Finally, we do not observe a significant difference in hyperparameter robustness between the single-task and multi-task models. When compared against the ResNet-50 single-task models, the HRNet-18 model seems to require less hyperparameter tuning to tackle the semantic segmentation and depth estimation tasks on NYUD-v2. We observe that the HRNet-18 model uses fewer parameters compared to the ResNet-50 model. This could explain why it is easier to train the single-task HRNet-18 models.  

\noindent\textbf{Conclusion.} We compared several decoder-focused architectures across two dense labeling datasets. On NYUD-v2, the performance on both semantic segmentation and depth estimation can be significantly improved by employing one of the decoder-focused models. However, both PAP-Net and JTRL incur a large increase in the number of FLOPS. MTI-Net and PAD-Net provide more viable alternatives when we wish to limit the number of FLOPS. On PASCAL, a multi-scale approach, like MTI-Net, seems better fitted for increasing the multi-scale performance while retaining the computational resources low. We conclude that the decoder-focused architectures obtain promising results on the MTL problem.

\subsection{Task Balancing}
\label{subsec: experiments_task_balancing}
We revisit the task balancing strategies from Section~\ref{subsec: task_balancing}. 

\noindent\textbf{NYUD-v2.} Table~\ref{tab: nyud_optimization} shows the results when training the MTL baseline with a ResNet-50 backbone using different task balancing strategies. On NYUD-v2, optimizing the loss weights with a grid search procedure resulted in a uniform loss weighting scheme. Therefore, in this case, the use of fixed uniform weights and the use of fixed weights from a grid search overlap. 

The MTL baseline with fixed weights improves over the single-task networks ($+0.41\%$ for $\Delta_{MTL}$). GradNorm can further boost the performance by adjusting the task-specific weights in the loss during training ($+1.45\%$). We conclude that uniform weights are suboptimal for training a multi-task model when the tasks employ different loss functions. 

Similar to GradNorm, DWA tries to balance the pace at which tasks are learned, but does not equalize the gradient magnitudes. From Table~\ref{tab: nyud_optimization}, we conclude that the latter is important as well ($-0.28\%$ with DWA vs $+1.45\%$ with GradNorm). Uncertainty weighting results in reduced performance compared to GradNorm ($-0.23\%$ vs $1.45\%$). Uncertainty weighting assigns a smaller weight to noisy or difficult tasks. Since the annotations on NYUD-v2 were not distilled, the noise levels are rather small. When we have access to clean ground-truth annotations, it seems better to balance the learning of tasks rather than lower the weights for the difficult tasks. Further, MGDA reports lower performance compared to GradNorm ($+0.02\%$ vs $+1.45\%$). MGDA only updates the weights along directions that are common among all task-specific gradients. The results suggest that it is better to allow some competition between the task-specific gradients in the shared layers, as this could help to avoid local minima. 

Finally, we conclude that the use of fixed loss weights optimized with a grid search still outperforms several existing task balancing methods. In particular, the solution obtained with fixed uniform weights outperforms the models trained with uncertainty weighting, MGDA and DWA on both the semantic segmentation and depth estimation tasks.  

\noindent\textbf{PASCAL.} The task balancing techniques are compared on PASCAL in Table~\ref{tab: pascal_optimization}. We find that a grid-search on the weight space works better than the use of automated task balancing procedures from Section~\ref{sec: optimization}. A similar observation was made by~\cite{maninis2019attentive}. We hypothesize that this is due to the imbalance between the optimal parameters, e.g. the weight for the edge detection loss is 100 times higher than for the semantic segmentation loss. 
Uncertainty weighting reports the highest performance losses on the distilled tasks, i.e. normals and saliency. This is because uncertainty weighting assigns a smaller weight to tasks with higher homoscedastic uncertainty. Differently, MGDA fails to properly learn the edge detection task. This is another indication (cf. NYUD-v2 results) that avoiding competing gradients by only back propagating along a common direction in the shared layers does not necessarily improve performance. This is particularly the case for the edge detection task on PASCAL because of two reasons. First, the edge detection loss has a much smaller magnitude compared to the other tasks. Second, when the edge annotations are converted into segments they provide an over-segmentation of the image. As a consequence, the loss gradient of the edge detection task often conflicts with other tasks since they have a smoother gradient. Because of this, MGDA prefers to mask out the gradient from the edge detection task by assigning it a smaller weight. Finally, GradNorm does not report higher performance when compared against uniform weighting. We hypothesize that this is due to the re-normalization of the loss weights during the optimization. The latter does not work well when the optimal loss magnitudes are very imbalanced. 

\noindent\textbf{Hyperparameters.} We use the performance profile from Figure~\ref{fig: hyperparam_optimization} to analyze the hyperparameter sensitivity of the task balancing methods on NYUD-v2. The used hyperparameters are the same as defined before (see Table~\ref{tab: grid_search}). GradNorm is the only technique that outperforms training a separate model for each task. However, when less effort is spent to tune the hyperparameters of the single-task models, all task balancing techniques result in improved performance. This observation explains the increased performance relative to the single-task case reported by prior works~\cite{kendall2018multi,chen2018gradnorm,sener2018multi}. Again, we stress the importance of carefully training the single-task baselines.

Second, as before, the MTL models seem more robust to hyperparameter changes compared to the single-task networks. In particular, the performance drops slower when using less optimal hyperparameter settings in the MTL case. 

Finally, fixed loss weighting and uncertainty weighting are most robust against hyperparameter changes. These techniques report high performance for a large number of hyperparameter settings. For example, out of the top-20\% best models, 40\% of those were trained with uncertainty weighting. 

\noindent\textbf{Discussion.} We evaluated the task balancing strategies from Section~\ref{subsec: task_balancing} under different settings. The methods were compared to selecting the loss weights by a grid-search procedure. Surprisingly, in our case, we found that grid-search is competitive or better compared to existing task balancing techniques. Also, a number of techniques performed worse than anticipated. Gong et al.~\cite{gong2019comparison} obtained similar results with ours, albeit only for a few loss balancing strategies. Also, Maninis et al.~\cite{maninis2019attentive} found that performing a grid search for the weights could outperform a state-of-the-art loss balancing scheme. Based on these works and our own findings, we argue that the optimization in MTL could benefit from further investigation (see also Section~\ref{subsubsec: task_balancing_discussion}).

Still existing task balancing techniques can be useful to train MTL networks. A grid search on the weight space becomes intractable when tackling large number of tasks. In this case, we can fallback to existing task balancing techniques to set the task weights. Furthermore, when dealing with noisy annotations, uncertainty weighting can help to automatically readjust the weights on the noisy tasks.

\subsection{Limitations}
We performed an extensive comparison of different MTL architectures and optimization strategies. This led to a number of important observations for each group of works. Undoubtedly, there are several aspects of the problem that fall outside the scope of our experimental analysis. We briefly discuss some of these limitations below.

\noindent\textbf{Joint architecture and optimization exploration.} The architectures and optimization strategies have been studied orthogonal to each other. However, the interaction between these two design spaces could be further examined in order to construct MTL systems at the level of a joint architecture-optimizer space.

\noindent\textbf{Influence data.} The tasks and training examples themselves play an important role in the optimization, e.g. uncertainty weighting works well when there are tasks with higher homoscedastic uncertainty. Thus, it would be useful to analyze what properties of the data are most important for the MTL setup. For example, such an analysis could be used to tune the optimization scheme to the needs of a specific setting.

\noindent\textbf{Hyperparameter robustness.} The performance profiles showed that the results strongly depend on the used hyperparameters. Interestingly, we found that some MTL algorithms are more robust against changes in the hyperparameter settings than others. This behavior could be further investigated.

\section{Related Domains}
\label{sec: related_domains}

So far, we focused on the application of MTL to jointly solve multiple vision tasks under a fully-supervised setting. In this section, we consider the MTL setup from a more general point-of-view, and analyze its connection with several related domains. The latter could potentially be combined with the MTL setup and improve it, or vice versa. 

\subsection{Multi-Domain Learning}
The methods considered so far were applied to solve multiple tasks on the same visual domain. However, there is a growing interest in learning representations that perform well for many visual domains simultaneously. For example, Bilen and Vedaldi~\cite{bilen2017universal} learned a compact multi-domain representation using domain-specific scaling parameters. This idea was later extended to the use of residual adapters~\cite{rebuffi2017learning,rebuffi2018efficient}. These works only explored multi-domain learning for various classification tasks. Future research should address the problem when considering multiple dense prediction tasks too. 

\subsection{Transfer Learning}
\textbf{Transfer learning}~\cite{pan2010survey} makes use of the knowledge obtained when solving one task, and applies it to tackle another task. Different from MTL, transfer learning does not consider solving all tasks concurrently. An important question in both transfer learning and MTL is whether visual tasks have relationships, i.e. they are correlated. Ma et al.~\cite{ma2018modeling} modeled the task relationships in a MTL network through a Multi-gate Mixture-of-Experts model. Zamir et al.~\cite{zamir2018taskonomy} provided a taxonomy for task transfer learning to quantify the task relationships. Similarly, Dwivedi and Roig~\cite{dwivedi2019representation} used Representation Similarity Analysis to obtain a measure of task affinity, by computing correlations between models pretrained on different tasks. Vandenhende et al.~\cite{vandenhende2019branched} then used these task relationships for the construction of a branched MTL network. Standley et al.~\cite{standley2019tasks} also relied on task relationships to define what tasks should be learned together in a MTL setup. 

\subsection{Neural Architecture Search}
The experimental results in Section~\ref{sec: experiments} showed that the success of MTL strongly depends on the use of a proper network architecture. Typically, such architectures are hand-crafted by human experts. However, given the size and complexity of the problem, this manual architecture exploration likely exceeds human design abilities. To automate the construction of the network architecture, \textbf{Neural Architecture Search} (NAS)~\cite{elsken2019neural} has been proposed in the literature. Yet, most existing NAS works are limited to task-specific models~\cite{zoph2017neural,liu2018progressive,pham2018efficient,liu2018darts,real2019regularized}. This is to be expected as using NAS for MTL assumes that layer sharing has to be jointly optimized with the layers types, their connectivity, etc., rendering the problem considerably expensive. 

To alleviate the heavy computational burden associated with NAS, several works have proposed to start from a predefined backbone network for which a cross-task layer sharing scheme is automatically determined. For example, Liang et al.~\cite{liang2018evolutionary} implemented an evolutionary architecture search for MTL, while others explored alternatives like branched MTL networks~\cite{lu2017fully,vandenhende2019branched,bruggemann2020automated,guo2020learning}, routing~\cite{rosenbaum2018routing}, stochastic filter grouping~\cite{bragman2019stochastic}, and feature partitioning~\cite{newell2019feature}. So far, NAS for MTL focused on how to share features across tasks in the encoder. We hypothesize that NAS could also be applied for the discovery of decoder-focused MTL models. 

\subsection{Other}
MTL has been applied to other problems too. This includes various domains, such as \textbf{language}~\cite{dong2015multi,mccann2018natural,liu2017adversarial}, \textbf{audio}~\cite{semanticbinaural}, \textbf{video}~\cite{diba2019holistic,pasunuru2017multi}, and \textbf{robotics}~\cite{wulfmeier2019regularized,hausman2018learning}, as well as with different learning paradigms, such as \textbf{reinforcement learning}~\cite{espeholt2018impala,wilson2007multi}, \textbf{self-supervised learning}~\cite{doersch2017multi}, \textbf{semi-supervised learning}~\cite{liu2008semi,zhang2009semi} and \textbf{active learning}~\cite{acharya2014active,reichart2008multi}. Surprisingly, in the deep learning era, very few works have considered MTL under the semi-supervised or active learning setting. Nonetheless, we believe that these are interesting directions for future research. 

For example, a major limitation of the fully-supervised MTL setting that we consider here is the requirement for all samples to be annotated for every task. Prior work~\cite{nekrasov2019real} showed that the standard update rule from Equation~\ref{eq: update_rule} gives suboptimal results if we do not take precautions when annotations are missing. To alleviate this problem, Kim et al.~\cite{kim2018disjoint} proposed to use an alternate learning scheme by updating the network weights for a single task at a time. A knowledge distillation term~\cite{hinton2015distilling} is included, in order to avoid loosing relevant information about the other tasks. Differently, Nekrasov et al.~\cite{nekrasov2019real} proposed to use the predictions from an expert model as synthetic ground-truth when annotations are missing. Although, these early attempts have shown encouraging results, we believe that this problem could benefit from further investigation. 

Finally, multi-task learning was recently shown to improve robustness. For example, in~\cite{maomultitask} a multi-task learning strategy showed robustness against adversarial attacks, while~\cite{zamir2020robust} found that applying cross-task consistency in MTL improves generalization, and allows for domain shift detection. 

\section{Conclusion}
\label{sec: conclusion}

In this paper, we reviewed recent methods for MTL within the scope of deep neural networks. First, we presented an extensive overview of both architectural and optimization based strategies for MTL. For each method, we described its key aspects, discussed the commonalities and differences with related works, and presented the possible advantages or disadvantages. Finally, we conducted an extensive experimental analysis of the described methods that led us to several key findings. We summarize some of our conclusions below, and present some possibilities for future work. 

First, the performance of MTL strongly varies depending on the task dictionary. Its size, task types, label sources, etc., all affect the final outcome. As a result, an appropriate architecture and optimization strategy should better be selected on a per case basis. Although we provided concrete observations as to why some methods work better for specific setups, MTL could generally benefit from a deeper theoretical understanding to maximize the expected gains in every case. For example, these gains seem dependent on a plurality of factors, e.g. amount of data, task relationships, noise, etc. Future work should try to isolate and analyze the influence of these different elements.  

Second, when it comes to tackling multiple dense prediction tasks using a single MTL model, decoder-focused architectures currently offer more advantages in terms of multi-task performance, and with limited computational overhead compared to the encoder-focused ones. As explained, this is due to an alignment of common cross-task patterns that decoder-focused architectures promote, which naturally fits well with dense prediction tasks. Encoder-focused architectures still offer certain advantages within the dense prediction task setting, but their inherent layer sharing seems better fitted to tackle multiple classification tasks. 

Finally, we analyzed multiple task balancing strategies, and isolated the elements that are most effective for balancing task learning, e.g. down-weighing noisy tasks, balancing task gradients, etc. Yet, many optimization aspects still remain poorly understood. For example, opposed to recent works, our analysis indicates that avoiding gradient competition between tasks can hurt performance. Furthermore, our study revealed that some task balancing strategies still suffer from shortcomings, and highlighted several discrepancies between existing methods. We hope that this work stimulates further research efforts into this problem. 

\noindent\textbf{Acknowledgment.} The authors would like to acknowledge support by Toyota via the TRACE project and MACCHINA (KU Leuven, C14/18/065). This work is also sponsored by the Flemish Government under the Flemish AI programme. Finally, the authors would like to thank Shikun Liu, Wanli Ouyang and the anonymous reviewers for useful feedback. 

\begin{IEEEbiography}[{\includegraphics[width=1in,height=1.25in,clip,keepaspectratio]{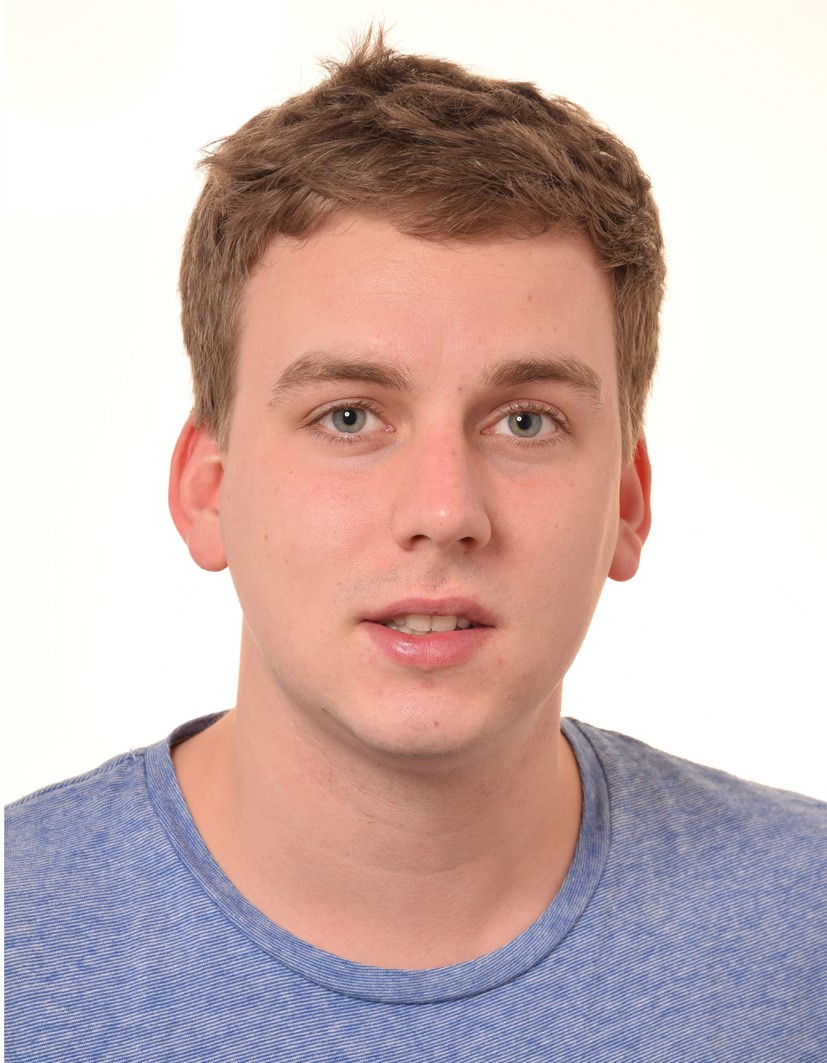}}]{Simon Vandenhende} received his MSE degree in Electrical Engineering from the KU Leuven, Belgium, in 2018. He is currently studying towards a PhD degree at the Center for Processing Speech and Images at KU Leuven. His research focuses on multi-task learning and self-supervised learning. He won a best paper award at MVA'19. He co-organized the first Commands For Autonomous Vehicles workshop at ECCV'20.
\end{IEEEbiography}

\begin{IEEEbiography}[{\includegraphics[width=1in,height=1.25in,clip,keepaspectratio]{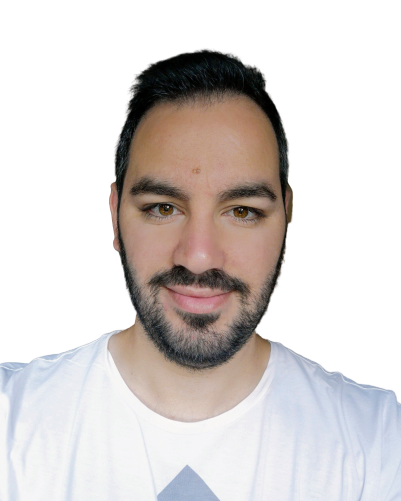}}]{Stamatios Georgoulis} is a post-doctoral researcher at the Computer Vision Lab of ETH Zurich. His current research interests include multi-task learning, unsupervised learning, and image generation. Before coming to Zurich, he was a doctoral student at the PSI group of KU Leuven, where he received his PhD under the supervision of Prof. Van Gool and Prof. Tuytelaars, focusing on extracting surface characteristics and lighting from images. Further back, he received his diploma in Electrical and Computer Engineering from the Aristotle University of Thessaloniki and participated in Microsoft's Imagine Cup Software Design Competition. He regularly serves as a reviewer in Computer Vision and Machine Learning conferences (CVPR, NeurIPS, ICCV, ECCV, etc.) with distinctions.
\end{IEEEbiography}

\begin{IEEEbiography}[{\includegraphics[width=1in,height=1.25in,clip,keepaspectratio]{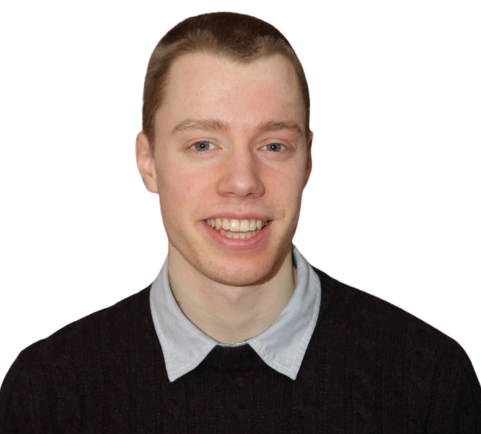}}]{Wouter Van Gansbeke} obtained his MSE degree in Electrical Engineering, with a focus on Information Technology, from the KU Leuven, Belgium, in 2018. He is currently pursuing a PhD degree at the Center for Processing Speech and Images at KU Leuven. His main research interests are self-supervised learning and multi-task learning. His focus is leveraging visual similarities, temporal information and multiple tasks to learn rich representations with limited annotations.
\end{IEEEbiography}

\begin{IEEEbiography}[{\includegraphics[width=1in,height=1.25in,clip,keepaspectratio]{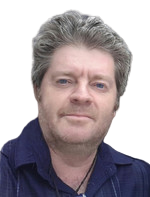}}]{Marc Proesmans} is a research associate at the Center for Processing Speech and Images at KU Leuven, where he leads the TRACE team. He has an extensive research experience on various aspects in image processing e.g. early vision processes, 3D reconstruction, optical flow, stereo, structure from motion, recognition. He has been involved in several European research projects, such as VANGUARD, IMPROOFS, MESH, MURALE, EUROPA, ROVINA. From 2000-2011, he has been CTO for a KU Leuven spin-off company specialized in special effects for the movie and game industry. He won several  prizes, Golden Egg, Tech-Art, a European IST, tech. Oscar nomination, European seal of Excellence. From 2015 he is leading the research on automotive driving as Research expert, and co-founded TRACE vzw to drive the technology transfer from academia to the real R\&D environment for autonomous driving in cooperation with Toyota. 
\end{IEEEbiography}

\begin{IEEEbiography}[{\includegraphics[width=1in,height=1.25in,clip,keepaspectratio]{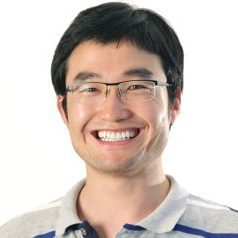}}]{Dengxin Dai} is a Senior Scientist working with the Computer Vision Lab at ETH Zurich. In 2016, he obtained his PhD in Computer Vision at ETH Zurich. Since then he is the Team Leader of TRACE-Zurich, working on Autonomous Driving within the R\&D project "TRACE: Toyota Research on Automated Cars in Europe". His research interests lie in autonomous driving, robust perception in adverse weather and illumination conditions, automotive sensors and computer vision under limited supervision. He has organized a CVPR Workshop series ('19, '20) on Vision for All Seasons: Bad Weather and Nighttime, and has organized an ICCV'19 workshop on Autonomous Driving. He has been a program committee member of several major computer vision conferences and received multiple outstanding reviewer awards. He is guest editor for the IJCV special issue Vision for All Seasons, area chair for WACV'20 and CVPR'21.
\end{IEEEbiography}

\begin{IEEEbiography}[{\includegraphics[width=1in,height=1.25in,clip,keepaspectratio]{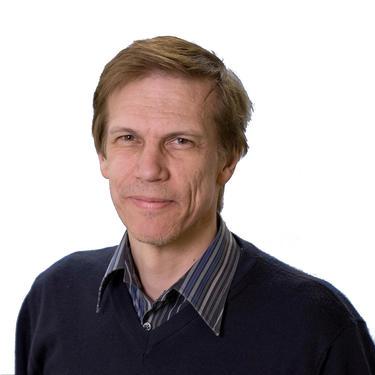}}]{Luc Van Gool}
is a full professor for Computer Vision at ETH Zurich and
the KU Leuven. He leads research and teaches at both places. He has authored over 300 papers. Luc Van Gool has been a program committee member of several, major
computer vision conferences (e.g. Program Chair ICCV'05, Beijing, General Chair of
ICCV'11, Barcelona, and of ECCV'14, Zurich). His main interests include 3D reconstruction and modeling, object recognition, and autonomous driving. He received several Best Paper awards (eg. David Marr Prize '98, Best Paper CVPR'07). He received the Koenderink Award in 2016 and `Distinguished Researcher' nomination by the IEEE Computer Society in 2017. In 2015 he also received they 5-yearly Excellence Prize by the Flemish Fund for Scientific Research. He was the holder of an ERC Advanced Grant (VarCity). Currently, he leads computer vision research for autonomous driving in the context of the Toyota TRACE labs in Leuven and at ETH, and has an extensive collaboration with Huawei on the issue of image and video enhancement.
\end{IEEEbiography}

\bibliographystyle{IEEEtran}
\bibliography{bibliography}

\begin{thebibliography}{100}
\providecommand{\url}[1]{#1}
\csname url@samestyle\endcsname
\providecommand{\newblock}{\relax}
\providecommand{\bibinfo}[2]{#2}
\providecommand{\BIBentrySTDinterwordspacing}{\spaceskip=0pt\relax}
\providecommand{\BIBentryALTinterwordstretchfactor}{4}
\providecommand{\BIBentryALTinterwordspacing}{\spaceskip=\fontdimen2\font plus
\BIBentryALTinterwordstretchfactor\fontdimen3\font minus
  \fontdimen4\font\relax}
\providecommand{\BIBforeignlanguage}[2]{{%
\expandafter\ifx\csname l@#1\endcsname\relax
\typeout{** WARNING: IEEEtran.bst: No hyphenation pattern has been}%
\typeout{** loaded for the language `#1'. Using the pattern for}%
\typeout{** the default language instead.}%
\else
\language=\csname l@#1\endcsname
\fi
#2}}
\providecommand{\BIBdecl}{\relax}
\BIBdecl

\bibitem{long2015fully}
J.~Long, E.~Shelhamer, and T.~Darrell, ``Fully convolutional networks for
  semantic segmentation,'' in \emph{CVPR}, 2015.

\bibitem{he2017mask}
K.~He, G.~Gkioxari, P.~Doll{\'a}r, and R.~Girshick, ``Mask r-cnn,'' in
  \emph{ICCV}, 2017.

\bibitem{eigen2014depth}
D.~Eigen, C.~Puhrsch, and R.~Fergus, ``Depth map prediction from a single image
  using a multi-scale deep network,'' in \emph{NIPS}, 2014.

\bibitem{gur2007direction}
M.~Gur and D.~M. Snodderly, ``Direction selectivity in v1 of alert monkeys:
  evidence for parallel pathways for motion processing,'' \emph{The Journal of
  physiology}, 2007.

\bibitem{misra2016cross}
I.~Misra, A.~Shrivastava, A.~Gupta, and M.~Hebert, ``Cross-stitch networks for
  multi-task learning,'' in \emph{CVPR}, 2016.

\bibitem{ruder2019latent}
S.~Ruder, J.~Bingel, I.~Augenstein, and A.~S{\o}gaard, ``Latent multi-task
  architecture learning,'' in \emph{AAAI}, 2019.

\bibitem{gao2019nddr}
Y.~Gao, J.~Ma, M.~Zhao, W.~Liu, and A.~L. Yuille, ``Nddr-cnn: Layerwise feature
  fusing in multi-task cnns by neural discriminative dimensionality
  reduction,'' in \emph{CVPR}, 2019.

\bibitem{liu2019end}
S.~Liu, E.~Johns, and A.~J. Davison, ``End-to-end multi-task learning with
  attention,'' in \emph{CVPR}, 2019.

\bibitem{lu2017fully}
Y.~Lu, A.~Kumar, S.~Zhai, Y.~Cheng, T.~Javidi, and R.~Feris, ``Fully-adaptive
  feature sharing in multi-task networks with applications in person attribute
  classification,'' in \emph{CVPR}, 2017.

\bibitem{vandenhende2019branched}
S.~Vandenhende, S.~Georgoulis, B.~De~Brabandere, and L.~Van~Gool, ``Branched
  multi-task networks: Deciding what layers to share,'' in \emph{BMVC}, 2020.

\bibitem{bruggemann2020automated}
D.~Bruggemann, M.~Kanakis, S.~Georgoulis, and L.~V. Gool, ``Automated search
  for resource-efficient branched multi-task networks,'' 2020.

\bibitem{guo2020learning}
P.~Guo, C.-Y. Lee, and D.~Ulbricht, ``Learning to branch for multi-task
  learning,'' in \emph{ICML}, 2020.

\bibitem{xu2018pad}
D.~Xu, W.~Ouyang, X.~Wang, and N.~Sebe, ``Pad-net: Multi-tasks guided
  prediction-and-distillation network for simultaneous depth estimation and
  scene parsing,'' in \emph{CVPR}, 2018.

\bibitem{zhang2019pattern}
Z.~Zhang, Z.~Cui, C.~Xu, Y.~Yan, N.~Sebe, and J.~Yang, ``Pattern-affinitive
  propagation across depth, surface normal and semantic segmentation,'' in
  \emph{CVPR}, 2019.

\bibitem{zhang2018joint}
Z.~Zhang, Z.~Cui, C.~Xu, Z.~Jie, X.~Li, and J.~Yang, ``Joint task-recursive
  learning for semantic segmentation and depth estimation,'' in \emph{ECCV},
  2018.

\bibitem{vandenhende2020mti}
S.~Vandenhende, S.~Georgoulis, and L.~Van~Gool, ``Mti-net: Multi-scale task
  interaction networks for multi-task learning,'' in \emph{ECCV}, 2020.

\bibitem{zhou2020pattern}
L.~Zhou, Z.~Cui, C.~Xu, Z.~Zhang, C.~Wang, T.~Zhang, and J.~Yang,
  ``Pattern-structure diffusion for multi-task learning,'' in \emph{CVPR},
  2020.

\bibitem{maninis2019attentive}
K.-K. Maninis, I.~Radosavovic, and I.~Kokkinos, ``Attentive single-tasking of
  multiple tasks,'' in \emph{CVPR}, 2019.

\bibitem{kendall2018multi}
A.~Kendall, Y.~Gal, and R.~Cipolla, ``Multi-task learning using uncertainty to
  weigh losses for scene geometry and semantics,'' in \emph{CVPR}, 2018.

\bibitem{chen2018gradnorm}
Z.~Chen, V.~Badrinarayanan, C.-Y. Lee, and A.~Rabinovich, ``Gradnorm: Gradient
  normalization for adaptive loss balancing in deep multitask networks,'' in
  \emph{ICML}, 2018.

\bibitem{guo2018dynamic}
M.~Guo, A.~Haque, D.-A. Huang, S.~Yeung, and L.~Fei-Fei, ``Dynamic task
  prioritization for multitask learning,'' in \emph{ECCV}, 2018.

\bibitem{sener2018multi}
O.~Sener and V.~Koltun, ``Multi-task learning as multi-objective
  optimization,'' in \emph{NIPS}, 2018.

\bibitem{lin2019pareto}
X.~Lin, H.-L. Zhen, Z.~Li, Q.-F. Zhang, and S.~Kwong, ``Pareto multi-task
  learning,'' in \emph{NIPS}, 2019.

\bibitem{liu2017adversarial}
P.~Liu, X.~Qiu, and X.~Huang, ``Adversarial multi-task learning for text
  classification,'' in \emph{ACL}, 2017.

\bibitem{sinha2018gradient}
A.~Sinha, Z.~Chen, V.~Badrinarayanan, and A.~Rabinovich, ``Gradient adversarial
  training of neural networks,'' \emph{arXiv preprint arXiv:1806.08028}, 2018.

\bibitem{zhao2018modulation}
X.~Zhao, H.~Li, X.~Shen, X.~Liang, and Y.~Wu, ``A modulation module for
  multi-task learning with applications in image retrieval,'' in \emph{ECCV},
  2018.

\bibitem{sanh2019hierarchical}
V.~Sanh, T.~Wolf, and S.~Ruder, ``A hierarchical multi-task approach for
  learning embeddings from semantic tasks,'' in \emph{AAAI}, 2019.

\bibitem{raffel2019exploring}
C.~Raffel, N.~Shazeer, A.~Roberts, K.~Lee, S.~Narang, M.~Matena, Y.~Zhou,
  W.~Li, and P.~J. Liu, ``Exploring the limits of transfer learning with a
  unified text-to-text transformer,'' \emph{arXiv preprint arXiv:1910.10683},
  2019.

\bibitem{chen2020just}
Z.~Chen, J.~Ngiam, Y.~Huang, T.~Luong, H.~Kretzschmar, Y.~Chai, and
  D.~Anguelov, ``Just pick a sign: Optimizing deep multitask models with
  gradient sign dropout,'' \emph{NeurIPS}, 2020.

\bibitem{caruana1997multitask}
R.~Caruana, ``Multitask learning,'' \emph{Machine learning}, 1997.

\bibitem{zhang2017survey}
Y.~Zhang and Q.~Yang, ``A survey on multi-task learning,'' \emph{arXiv preprint
  arXiv:1707.08114}, 2017.

\bibitem{wang2018glue}
A.~Wang, A.~Singh, J.~Michael, F.~Hill, O.~Levy, and S.~Bowman, ``Glue: A
  multi-task benchmark and analysis platform for natural language
  understanding,'' in \emph{Workshop EMNLP}, 2018.

\bibitem{deng2013new}
L.~Deng, G.~Hinton, and B.~Kingsbury, ``New types of deep neural network
  learning for speech recognition and related applications: An overview,'' in
  \emph{2013 IEEE international conference on acoustics, speech and signal
  processing}, 2013.

\bibitem{widmer2010leveraging}
C.~Widmer, J.~Leiva, Y.~Altun, and G.~R{\"a}tsch, ``Leveraging sequence
  classification by taxonomy-based multitask learning,'' in \emph{Annual
  International Conference on Research in Computational Molecular Biology},
  2010.

\bibitem{bell2020groknet}
S.~Bell, Y.~Liu, S.~Alsheikh, Y.~Tang, E.~Pizzi, M.~Henning, K.~Singh,
  O.~Parkhi, and F.~Borisyuk, ``Groknet: Unified computer vision model trunk
  and embeddings for commerce,'' in \emph{SIGKDD}, 2020.

\bibitem{ruder2017overview}
S.~Ruder, ``An overview of multi-task learning in deep neural networks,''
  \emph{arXiv preprint arXiv:1706.05098}, 2017.

\bibitem{gong2019comparison}
T.~Gong, T.~Lee, C.~Stephenson, V.~Renduchintala, S.~Padhy, A.~Ndirango,
  G.~Keskin, and O.~H. Elibol, ``A comparison of loss weighting strategies for
  multi task learning in deep neural networks,'' \emph{IEEE Access}, 2019.

\bibitem{evgeniou2004regularized}
T.~Evgeniou and M.~Pontil, ``Regularized multi--task learning,'' in \emph{KDD},
  2004.

\bibitem{xue2007multi}
Y.~Xue, X.~Liao, L.~Carin, and B.~Krishnapuram, ``Multi-task learning for
  classification with dirichlet process priors,'' \emph{JMLR}, 2007.

\bibitem{jacob2009clustered}
L.~Jacob, J.-p. Vert, and F.~R. Bach, ``Clustered multi-task learning: A convex
  formulation,'' in \emph{NIPS}, 2009.

\bibitem{zhou2011clustered}
J.~Zhou, J.~Chen, and J.~Ye, ``Clustered multi-task learning via alternating
  structure optimization,'' in \emph{NIPS}, 2011.

\bibitem{bakker2003task}
B.~Bakker and T.~Heskes, ``Task clustering and gating for bayesian multitask
  learning,'' \emph{JMLR}, 2003.

\bibitem{yu2005learning}
K.~Yu, V.~Tresp, and A.~Schwaighofer, ``Learning gaussian processes from
  multiple tasks,'' in \emph{ICML}, 2005.

\bibitem{lee2007learning}
S.-I. Lee, V.~Chatalbashev, D.~Vickrey, and D.~Koller, ``Learning a meta-level
  prior for feature relevance from multiple related tasks,'' in \emph{ICML},
  2007.

\bibitem{daume2009bayesian}
H.~Daum{\'e}~III, ``Bayesian multitask learning with latent hierarchies,''
  \emph{arXiv preprint arXiv:0907.0783}, 2009.

\bibitem{kumar2012learning}
A.~Kumar and H.~Daume~III, ``Learning task grouping and overlap in multi-task
  learning,'' in \emph{ICML}, 2012.

\bibitem{argyriou2008convex}
A.~Argyriou, T.~Evgeniou, and M.~Pontil, ``Convex multi-task feature
  learning,'' \emph{Machine learning}, 2008.

\bibitem{liu2012multi}
J.~Liu, S.~Ji, and J.~Ye, ``Multi-task feature learning via efficient l2,
  1-norm minimization,'' in \emph{Uncertainty in Artificial Intelligence},
  2009.

\bibitem{jalali2010dirty}
A.~Jalali, S.~Sanghavi, C.~Ruan, and P.~K. Ravikumar, ``A dirty model for
  multi-task learning,'' in \emph{NIPS}, 2010.

\bibitem{agarwal2010learning}
A.~Agarwal, S.~Gerber, and H.~Daume, ``Learning multiple tasks using manifold
  regularization,'' in \emph{NIPS}, 2010.

\bibitem{ando2005framework}
R.~K. Ando and T.~Zhang, ``A framework for learning predictive structures from
  multiple tasks and unlabeled data,'' \emph{JMLR}, 2005.

\bibitem{rai2010infinite}
P.~Rai and H.~Daum{\'e}~III, ``Infinite predictor subspace models for multitask
  learning,'' in \emph{AISTATS}, 2010.

\bibitem{neven2017fast}
D.~Neven, B.~De~Brabandere, S.~Georgoulis, M.~Proesmans, and L.~Van~Gool,
  ``Fast scene understanding for autonomous driving,'' in \emph{IV Workshops},
  2017.

\bibitem{teichmann2018multinet}
M.~Teichmann, M.~Weber, M.~Zoellner, R.~Cipolla, and R.~Urtasun, ``Multinet:
  Real-time joint semantic reasoning for autonomous driving,'' in \emph{IV},
  2018.

\bibitem{kokkinos2017ubernet}
I.~Kokkinos, ``Ubernet: Training a universal convolutional neural network for
  low-, mid-, and high-level vision using diverse datasets and limited
  memory,'' in \emph{CVPR}, 2017.

\bibitem{long2017learning}
M.~Long, Z.~Cao, J.~Wang, and S.~Y. Philip, ``Learning multiple tasks with
  multilinear relationship networks,'' in \emph{NIPS}, 2017.

\bibitem{bragman2019stochastic}
F.~J. Bragman, R.~Tanno, S.~Ourselin, D.~C. Alexander, and M.~J. Cardoso,
  ``Stochastic filter groups for multi-task cnns: Learning specialist and
  generalist convolution kernels,'' in \emph{ICCV}, 2019.

\bibitem{he2015spatial}
K.~He, X.~Zhang, S.~Ren, and J.~Sun, ``Spatial pyramid pooling in deep
  convolutional networks for visual recognition,'' \emph{TPAMI}, 2015.

\bibitem{chen2017deeplab}
L.-C. Chen, G.~Papandreou, I.~Kokkinos, K.~Murphy, and A.~L. Yuille, ``Deeplab:
  Semantic image segmentation with deep convolutional nets, atrous convolution,
  and fully connected crfs,'' \emph{TPAMI}, 2017.

\bibitem{zhao2017pyramid}
H.~Zhao, J.~Shi, X.~Qi, X.~Wang, and J.~Jia, ``Pyramid scene parsing network,''
  in \emph{CVPR}, 2017.

\bibitem{yuan2018ocnet}
Y.~Yuan and J.~Wang, ``Ocnet: Object context network for scene parsing,''
  \emph{arXiv preprint arXiv:1809.00916}, 2018.

\bibitem{yosinski2014transferable}
J.~Yosinski, J.~Clune, Y.~Bengio, and H.~Lipson, ``How transferable are
  features in deep neural networks?'' in \emph{NIPS}, 2014.

\bibitem{dwivedi2019representation}
K.~Dwivedi and G.~Roig, ``Representation similarity analysis for efficient task
  taxonomy \& transfer learning,'' in \emph{CVPR}, 2019.

\bibitem{meyerson2018beyond}
E.~Meyerson and R.~Miikkulainen, ``Beyond shared hierarchies: Deep multitask
  learning through soft layer ordering,'' in \emph{ICLR}, 2018.

\bibitem{yang2016deep}
Y.~Yang and T.~Hospedales, ``Deep multi-task representation learning: A tensor
  factorisation approach,'' \emph{arXiv preprint arXiv:1605.06391}, 2016.

\bibitem{rosenbaum2018routing}
C.~Rosenbaum, T.~Klinger, and M.~Riemer, ``Routing networks: Adaptive selection
  of non-linear functions for multi-task learning,'' in \emph{ICLR}, 2018.

\bibitem{mallya2018piggyback}
A.~Mallya, D.~Davis, and S.~Lazebnik, ``Piggyback: Adapting a single network to
  multiple tasks by learning to mask weights,'' in \emph{ECCV}, 2018.

\bibitem{huang2018gnas}
S.~Huang, X.~Li, Z.~Cheng, A.~Hauptmann \emph{et~al.}, ``Gnas: A greedy neural
  architecture search method for multi-attribute learning,'' in \emph{ACMMM},
  2018.

\bibitem{newell2019feature}
A.~Newell, L.~Jiang, C.~Wang, L.-J. Li, and J.~Deng, ``Feature partitioning for
  efficient multi-task architectures,'' \emph{arXiv preprint arXiv:1908.04339},
  2019.

\bibitem{suteu2019regularizing}
M.~Suteu and Y.~Guo, ``Regularizing deep multi-task networks using orthogonal
  gradients,'' \emph{arXiv preprint arXiv:1912.06844}, 2019.

\bibitem{lin2017focal}
T.-Y. Lin, P.~Goyal, R.~Girshick, K.~He, and P.~Doll{\'a}r, ``Focal loss for
  dense object detection,'' in \emph{ICCV}, 2017.

\bibitem{desideri2012multiple}
J.-A. D{\'e}sid{\'e}ri, ``Multiple-gradient descent algorithm (mgda) for
  multiobjective optimization,'' \emph{Comptes Rendus Mathematique}, 2012.

\bibitem{silberman2012indoor}
N.~Silberman, D.~Hoiem, P.~Kohli, and R.~Fergus, ``Indoor segmentation and
  support inference from rgbd images,'' in \emph{ECCV}, 2012.

\bibitem{everingham2010pascal}
M.~Everingham, L.~Van~Gool, C.~K. Williams, J.~Winn, and A.~Zisserman, ``The
  pascal visual object classes (voc) challenge,'' \emph{IJCV}, 2010.

\bibitem{chen2014detect}
X.~Chen, R.~Mottaghi, X.~Liu, S.~Fidler, R.~Urtasun, and A.~Yuille, ``Detect
  what you can: Detecting and representing objects using holistic models and
  body parts,'' in \emph{CVPR}, 2014, pp. 1971--1978.

\bibitem{bansal2017pixelnet}
A.~Bansal, X.~Chen, B.~Russell, A.~Gupta, and D.~Ramanan, ``Pixelnet:
  Representation of the pixels, by the pixels, and for the pixels,''
  \emph{arXiv preprint arXiv:1702.06506}, 2017.

\bibitem{chen2018encoder}
L.-C. Chen, Y.~Zhu, G.~Papandreou, F.~Schroff, and H.~Adam, ``Encoder-decoder
  with atrous separable convolution for semantic image segmentation,'' in
  \emph{ECCV}, 2018.

\bibitem{martin2004learning}
D.~R. Martin, C.~C. Fowlkes, and J.~Malik, ``Learning to detect natural image
  boundaries using local brightness, color, and texture cues,'' \emph{TPAMI},
  2004.

\bibitem{he2016deep}
K.~He, X.~Zhang, S.~Ren, and J.~Sun, ``Deep residual learning for image
  recognition,'' in \emph{CVPR}, 2016.

\bibitem{sun2019deep}
K.~Sun, B.~Xiao, D.~Liu, and J.~Wang, ``Deep high-resolution representation
  learning for human pose estimation,'' in \emph{CVPR}, 2019.

\bibitem{zou2018df}
Y.~Zou, Z.~Luo, and J.-B. Huang, ``Df-net: Unsupervised joint learning of depth
  and flow using cross-task consistency,'' in \emph{ECCV}, 2018.

\bibitem{girshick2015fast}
R.~Girshick, ``Fast r-cnn,'' in \emph{ICCV}, 2015.

\bibitem{ren2015faster}
S.~Ren, K.~He, R.~Girshick, and J.~Sun, ``Faster r-cnn: Towards real-time
  object detection with region proposal networks,'' in \emph{NIPS}, 2015.

\bibitem{dvornik2017blitznet}
N.~Dvornik, K.~Shmelkov, J.~Mairal, and C.~Schmid, ``Blitznet: A real-time deep
  network for scene understanding,'' in \emph{ICCV}, 2017.

\bibitem{bilen2017universal}
H.~Bilen and A.~Vedaldi, ``Universal representations: The missing link between
  faces, text, planktons, and cat breeds,'' \emph{arXiv preprint
  arXiv:1701.07275}, 2017.

\bibitem{rebuffi2017learning}
S.-A. Rebuffi, H.~Bilen, and A.~Vedaldi, ``Learning multiple visual domains
  with residual adapters,'' in \emph{NIPS}, 2017.

\bibitem{rebuffi2018efficient}
------, ``Efficient parametrization of multi-domain deep neural networks,'' in
  \emph{CVPR}, 2018.

\bibitem{pan2010survey}
S.~J. Pan, Q.~Yang \emph{et~al.}, ``A survey on transfer learning,''
  \emph{TKDE}, 2010.

\bibitem{ma2018modeling}
J.~Ma, Z.~Zhao, X.~Yi, J.~Chen, L.~Hong, and E.~H. Chi, ``Modeling task
  relationships in multi-task learning with multi-gate mixture-of-experts,'' in
  \emph{KDD}, 2018.

\bibitem{zamir2018taskonomy}
A.~R. Zamir, A.~Sax, W.~Shen, L.~J. Guibas, J.~Malik, and S.~Savarese,
  ``Taskonomy: Disentangling task transfer learning,'' in \emph{CVPR}, 2018.

\bibitem{standley2019tasks}
T.~Standley, A.~R. Zamir, D.~Chen, L.~Guibas, J.~Malik, and S.~Savarese,
  ``Which tasks should be learned together in multi-task learning?''
  \emph{ICML}, 2020.

\bibitem{elsken2019neural}
T.~Elsken, J.~H. Metzen, and F.~Hutter, ``Neural architecture search: A
  survey.'' \emph{JMLR}, 2019.

\bibitem{zoph2017neural}
B.~Zoph and Q.~V. Le, ``Neural architecture search with reinforcement
  learning,'' in \emph{ICLR}, 2017.

\bibitem{liu2018progressive}
C.~Liu, B.~Zoph, M.~Neumann, J.~Shlens, W.~Hua, L.-J. Li, L.~Fei-Fei,
  A.~Yuille, J.~Huang, and K.~Murphy, ``Progressive neural architecture
  search,'' in \emph{ECCV}, 2018.

\bibitem{pham2018efficient}
H.~Pham, M.~Y. Guan, B.~Zoph, Q.~V. Le, and J.~Dean, ``Efficient neural
  architecture search via parameter sharing,'' in \emph{ICML}, 2018.

\bibitem{liu2018darts}
H.~Liu, K.~Simonyan, and Y.~Yang, ``Darts: Differentiable architecture
  search,'' \emph{ICLR}, 2018.

\bibitem{real2019regularized}
E.~Real, A.~Aggarwal, Y.~Huang, and Q.~V. Le, ``Regularized evolution for image
  classifier architecture search,'' in \emph{AAAI}, 2019.

\bibitem{liang2018evolutionary}
J.~Liang, E.~Meyerson, and R.~Miikkulainen, ``Evolutionary architecture search
  for deep multitask networks,'' in \emph{GECCO}, 2018.

\bibitem{dong2015multi}
D.~Dong, H.~Wu, W.~He, D.~Yu, and H.~Wang, ``Multi-task learning for multiple
  language translation,'' in \emph{ACL}, 2015.

\bibitem{mccann2018natural}
B.~McCann, N.~S. Keskar, C.~Xiong, and R.~Socher, ``The natural language
  decathlon: Multitask learning as question answering,'' \emph{arXiv preprint
  arXiv:1806.08730}, 2018.

\bibitem{semanticbinaural}
A.~{Balajee Vasudevan}, D.~Dai, and L.~{Van Gool}, ``Semantic object prediction
  and spatial sound prediction with binaural sounds,'' in \emph{ECCV}, 2020.

\bibitem{diba2019holistic}
A.~Diba, M.~Fayyaz, V.~Sharma, M.~Paluri, J.~Gall, R.~Stiefelhagen, and
  L.~Van~Gool, ``Holistic large scale video understanding,'' \emph{arXiv
  preprint arXiv:1904.11451}, 2019.

\bibitem{pasunuru2017multi}
R.~Pasunuru and M.~Bansal, ``Multi-task video captioning with video and
  entailment generation,'' in \emph{ACL}, 2017.

\bibitem{wulfmeier2019regularized}
M.~Wulfmeier, A.~Abdolmaleki, R.~Hafner, J.~T. Springenberg, M.~Neunert,
  T.~Hertweck, T.~Lampe, N.~Siegel, N.~Heess, and M.~Riedmiller, ``Regularized
  hierarchical policies for compositional transfer in robotics,'' \emph{arXiv
  preprint arXiv:1906.11228}, 2019.

\bibitem{hausman2018learning}
K.~Hausman, J.~T. Springenberg, Z.~Wang, N.~Heess, and M.~Riedmiller,
  ``Learning an embedding space for transferable robot skills,'' in
  \emph{ICLR}, 2018.

\bibitem{espeholt2018impala}
L.~Espeholt, H.~Soyer, R.~Munos, K.~Simonyan, V.~Mnih, T.~Ward, Y.~Doron,
  V.~Firoiu, T.~Harley, I.~Dunning \emph{et~al.}, ``Impala: Scalable
  distributed deep-rl with importance weighted actor-learner architectures,''
  in \emph{ICML}, 2018.

\bibitem{wilson2007multi}
A.~Wilson, A.~Fern, S.~Ray, and P.~Tadepalli, ``Multi-task reinforcement
  learning: a hierarchical bayesian approach,'' in \emph{ICML}, 2007.

\bibitem{doersch2017multi}
C.~Doersch and A.~Zisserman, ``Multi-task self-supervised visual learning,'' in
  \emph{ICCV}, 2017.

\bibitem{liu2008semi}
Q.~Liu, X.~Liao, and L.~Carin, ``Semi-supervised multitask learning,'' in
  \emph{NIPS}, 2008.

\bibitem{zhang2009semi}
Y.~Zhang and D.-Y. Yeung, ``Semi-supervised multi-task regression,'' in
  \emph{KDD}, 2009.

\bibitem{acharya2014active}
A.~Acharya, R.~J. Mooney, and J.~Ghosh, ``Active multitask learning using both
  latent and supervised shared topics,'' in \emph{ICDM}, 2014.

\bibitem{reichart2008multi}
R.~Reichart, K.~Tomanek, U.~Hahn, and A.~Rappoport, ``Multi-task active
  learning for linguistic annotations,'' in \emph{ACL}, 2008.

\bibitem{nekrasov2019real}
V.~Nekrasov, T.~Dharmasiri, A.~Spek, T.~Drummond, C.~Shen, and I.~Reid,
  ``Real-time joint semantic segmentation and depth estimation using asymmetric
  annotations,'' in \emph{ICRA}, 2019.

\bibitem{kim2018disjoint}
D.-J. Kim, J.~Choi, T.-H. Oh, Y.~Yoon, and I.~S. Kweon, ``Disjoint multi-task
  learning between heterogeneous human-centric tasks,'' in \emph{WACV}.\hskip
  1em plus 0.5em minus 0.4em\relax IEEE, 2018.

\bibitem{hinton2015distilling}
G.~Hinton, O.~Vinyals, and J.~Dean, ``Distilling the knowledge in a neural
  network,'' \emph{arXiv preprint arXiv:1503.02531}, 2015.

\bibitem{maomultitask}
C.~Mao, A.~Gupta, V.~Nitin, B.~Ray, S.~Song, J.~Yang, and C.~Vondrick,
  ``Multitask learning strengthens adversarial robustness,'' 2020.

\bibitem{zamir2020robust}
A.~R. Zamir, A.~Sax, N.~Cheerla, R.~Suri, Z.~Cao, J.~Malik, and L.~J. Guibas,
  ``Robust learning through cross-task consistency,'' in \emph{CVPR}, 2020.

\end{thebibliography}

\end{document}